\documentclass[11pt]{article}

\usepackage{amsmath}
\usepackage{amssymb}
\usepackage{array} 
\usepackage{booktabs}
\usepackage{multirow}
\usepackage{arydshln}

\usepackage[preprint]{acl}

\usepackage{times}
\usepackage{latexsym}

\usepackage[T1]{fontenc}

\usepackage[utf8]{inputenc}

\usepackage{microtype}

\usepackage{inconsolata}

\usepackage{graphicx}
\usepackage{subcaption}
\usepackage{titlesec}      
\usepackage{titletoc}      
\usepackage[most]{tcolorbox}
\usepackage{hyperref}
\definecolor{OliveGreen}{RGB}{40, 176, 77}
\definecolor{BrickRed}{rgb}{0.8, 0.25, 0.33}

\title{CheckRLM: Effective Knowledge-Thought Coherence Checking in Retrieval-Augmented Reasoning}

\author{
 Dingling~Xu$^{12*}$, Ruobing~Wang$^{34}$\thanks{Equal contribution.
 \texttt{202111081066@mail.bnu.edu.cn}},
 \textbf{Qingfei~Zhao}$^{34}$, \textbf{Yukun~Yan}$^{5\dag}$, \textbf{Zhichun~Wang}$^{12}\thanks{  Corresponding authors.  \texttt{yanyk.thu@gmail.com};
 \texttt{zcwang@bnu.edu.cn}}$, \\  \textbf{Daren~Zha$^{3}$,} 
 \textbf{Shi~Yu$^{5}$,}
 \textbf{Zhenghao~Liu$^{6}$,}   \textbf{Shuo~Wang$^5$,}
 \textbf{Xu~Han$^5$,}
 \textbf{Maosong Sun}$^{5}$ \\
 $^1$School of Artificial Intelligence, Beijing Normal University, Beijing 100875, China\\
 $^2$Beijing Key Laboratory of Artificial Intelligence for Education, Beijing 100875, China\\
 $^3$Institute of Information Engineering, Chinese Academy of Sciences,\\
 $^4$School of Cyber Security, University of Chinese Academy of Sciences,\\
 $^5$Department of Computer Science and Technology, Institute for AI, Tsinghua University,\\
 $^6$Northeastern University
}

\begin{document}
\maketitle
\begin{abstract}
Reasoning Language Models (RLMs) have significantly improved performance on complex tasks by extending the reasoning chain. However, these chains are prone to containing factual errors, particularly in knowledge-intensive tasks. To address this issue, we propose \textbf{CheckRLM}, a framework that improves the reliability of the reasoning process through Retrieval-Augmented Generation (RAG) by timely checking and correcting factual errors. Specifically, CheckRLM extracts factual claims from the reasoning chain to identify and localize subtle knowledge inconsistencies during inference. Upon detection of errors, a refinement mechanism performs minimal-cost yet precise corrections by leveraging external knowledge, ensuring coherence between the reasoning chain and correct knowledge. Extensive experiments demonstrate that CheckRLM substantially outperforms existing baselines, exhibiting a strong capability to mitigate error accumulation in long-horizon reasoning with lower costs.
The code and data are available at \url{https://github.com/AI9Stars/CheckRLM}.
\end{abstract}

\section{Introduction}
Reasoning Language Models (RLMs) such as OpenAI-o1 ~\cite{openai-o1} and DeepSeek-R1 ~\cite{guo_deepseek-r1_2025} have notable advanced reasoning capabilities with reinforcement learning and test-time scaling ~\cite{11267249,gao2025synergizing}. 
By decomposing complex problems and engaging in iterative exploration and reflection over long reasoning chains, RLMs attract significant attention in logic-intensive domains, including mathematics and coding ~\cite{chen2025towards,li2025survey}. 

Despite these advancements, the extended reasoning chain also introduces the critical challenge of maintaining factual coherence throughout the process, especially in knowledge-intensive tasks ~\cite{he2025can,yao2025reasoning}. In long stepwise reasoning, we observe that the factual accuracy of each intermediate step is paramount, as it forms the foundational premise for subsequent deductions. Consequently, any factual error at a given step can be propagated, leading to final answer deviations--a phenomenon termed error accumulation ~\cite{ling2023deductive,tyen2024llms}. 
\begin{figure}[t]
  \includegraphics[width=\columnwidth]{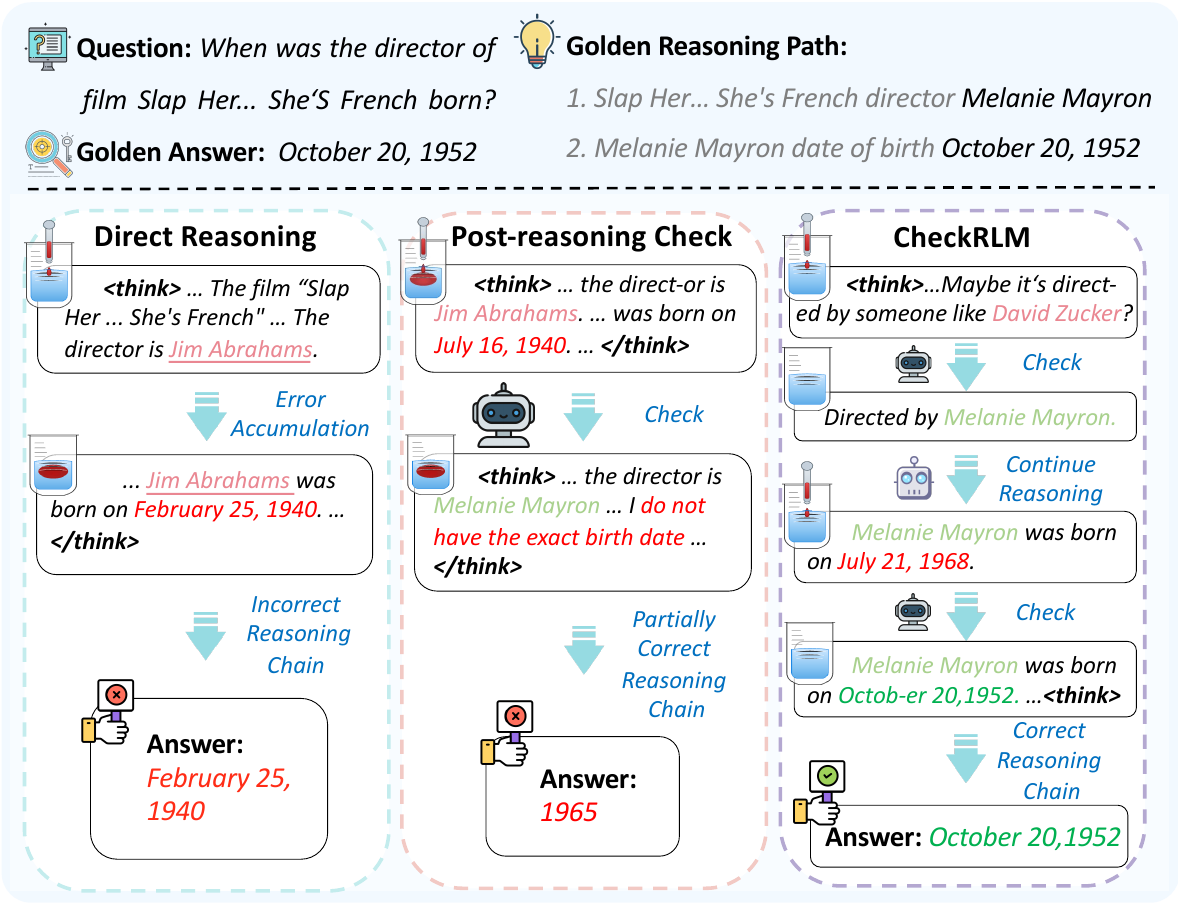}
  \caption{\textbf{Illustration of error accumulation and CheckRLM.}
Direct Reasoning and Post-reasoning Check suffer from erroneous internal knowledge in the RLM and insufficient external knowledge, whereas CheckRLM corrects errors timely to reach the correct answer.}
  \label{fig:example-fig1}
\end{figure}
\textbf{As illustrated in Figure~\ref{fig:example-fig1}},
Direct Reasoning initially produces an incorrect director name ``\textit{Jim Abrahams}'', which is then propagated as a premise for subsequent reasoning, leading to an erroneous inference of the director’s birth date and the final incorrect answer ``\textit{February 25, 1940}''. Although Post-reasoning Check later revises the director's name, it fails to repair the already corrupted intermediate reasoning.

To bridge this gap, we propose \textbf{CheckRLM}, a framework that employs Retrieval-Augmented Generation (RAG) to promptly identify and correct factual errors within long reasoning chains, thereby aligning them with external knowledge. To effectively check and refine reasoning chains, CheckRLM comprises two components: in-process knowledge claim recognition and localized knowledge coherence correction via retrieval. During in-process knowledge claim recognition, key factual claims pertinent to the query are identified and extracted from each newly generated reasoning segment, facilitating the location of potential inconsistencies while minimizing interference from irrelevant information. In the localized knowledge coherence correction via retrieval process, we use the extracted key factual claims along with the original question to search an external knowledge base and perform token-level precise corrections, ensuring that each segment of the verified reasoning chain aligns with external knowledge and reducing error accumulation.

In summary, our contributions are as follows: 1) we propose CheckRLM, an efficient RAG framework with in-reasoning knowledge checking that identifies and corrects factual errors during long-chain reasoning, effectively preventing error accumulation; 2) we introduce a localized intermediate intervention mechanism that dynamically extracts and verifies critical knowledge claims from partial reasoning states, enabling timely and fine-grained factual correction; 3) CheckRLM significantly outperforms recent strong baselines, while comprehensive analyses show that it enables early, targeted error correction and simultaneously achieves top-tier performance with significantly reduced inference time and token consumption.

\section{Related Work}
\subsection{Reasoning Language Model (RLM)}
RLMs enhance the quality of final answers by guiding Large Language Models (LLMs) to decompose complex tasks into multiple subquestions and by extending step-by-step reasoning chains ~\cite{chen2025towards,gao2025synergizing,11267249,li2025survey}. Prior work primarily relies on engineered prompts during inference without modifying the internal parameters of LLMs ~\cite{ICLR2024_4990dad2,zhou2023leasttomost,zhang2023automatic,yao2023tree}. 
Chain-of-Thought ~\cite{wei2022chain} inserts several task-oriented exemplars that contain intermediate reasoning steps into the prompt, while ~\cite{kojima2022large} stimulates zero-shot generation of intermediate reasoning by adding the instruction ``\textit{let's think step by step}'' to task instructions, effectively eliciting reasoning capabilities in LLMs. However, these prompt-engineering approaches are largely constrained by the quality of the prompts and exemplars, and the depth and correctness of the elicited reasoning are difficult to guarantee.
Recent studies ~\cite{openai-o1,guo_deepseek-r1_2025,yang2025qwen3} have focused on deep optimization of LLMs via large-scale reinforcement learning (RL). By designing verifiable reward functions, RLMs exhibit high-level reasoning behaviors, including self-verification, reflection, and dynamic strategy adjustment; furthermore, increasing test-time compute yields continued improvements in reasoning performance ~\cite{weng2023large,kumar2025training}. 
Meanwhile, concurrent studies have pointed out that these RLMs tend to produce hallucinations within their long reasoning chains ~\cite{he2025can,yao2025reasoning}. An early factual error can subsequently mislead the entire reasoning trajectory, leading to a phenomenon known as ``error accumulation'' ~\cite{ling2023deductive,tyen2024llms}.

\begin{figure*}[ht]
  \centering
\includegraphics[width=\textwidth]{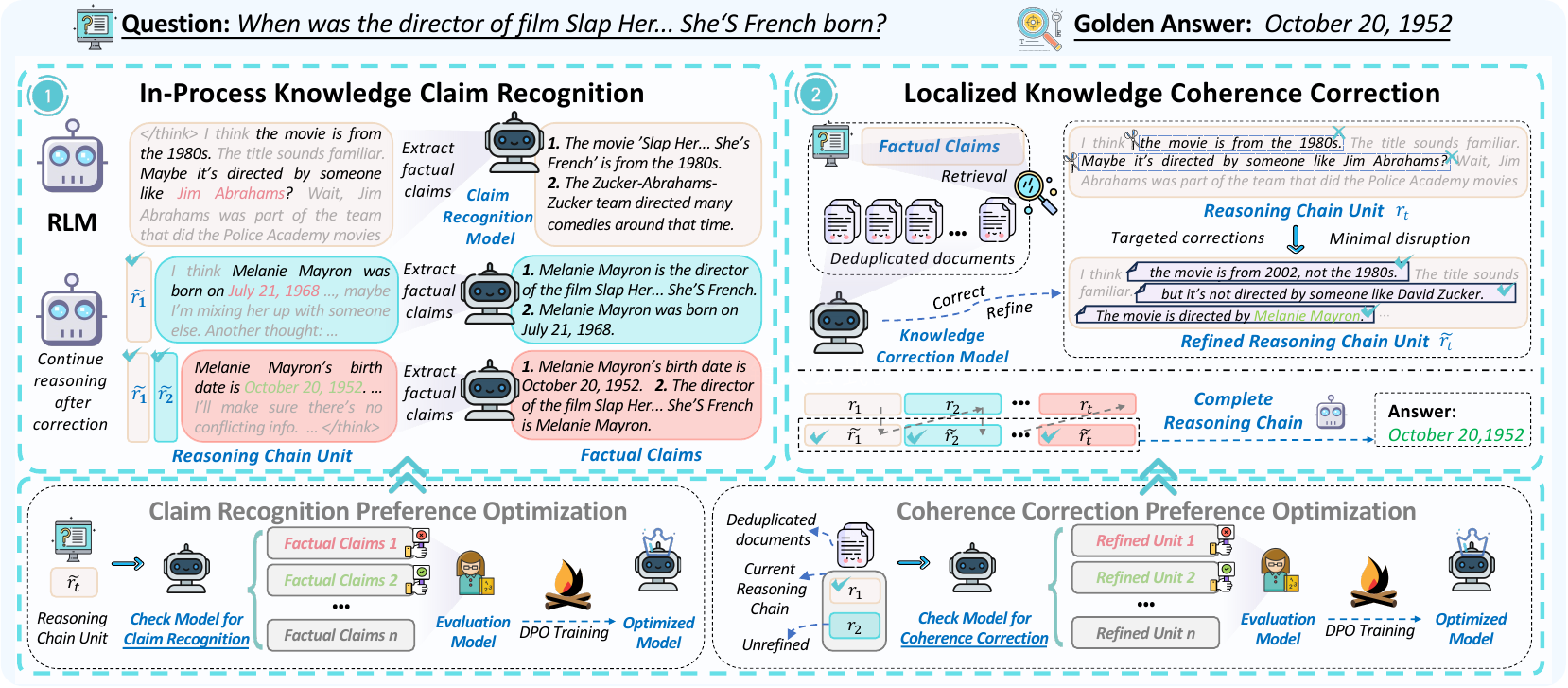}
  \caption{\textbf{Overview of CheckRLM.} In-process knowledge claim recognition locates error positions by extracting factual claims from reasoning chain units, while localized knowledge coherence correction corrects factual errors at minimal cost based on retrieved documents. The two modules are jointly optimized using DPO.}
  \label{fig2: Overview of K-COHERE}
\end{figure*}

\subsection{Retrieval-Augmented Generation (RAG)}
RAG effectively mitigates the ``hallucination'' in knowledge-intensive tasks by integrating LLMs with external knowledge bases, improving the accuracy and reliability of content ~\cite{lewis2020retrieval,ram2023context,zhao-etal-2024-longrag,zhu-etal-2025-rageval}. However, the single-retrieval mechanism in vanilla RAG ~\cite{izacard2023atlas} often yields limited information, struggling to cover the complete knowledge background required for complex questions ~\cite{wang-etal-2025-deepnote,wang2024searching}. To address the limitations of Vanilla RAG, researchers have proposed the adaptive RAG framework ~\cite{jeong-etal-2024-adaptive,jiang2023active,asai2024self,yao2022react}, which employs an intelligent decision-making mechanism to dynamically determine when to retrieve, what to retrieve, and how to integrate the retrieved information. This significantly improves both the response quality and efficiency of RAG systems. However, these traditional adaptive RAG methods are not suitable for RLMs based on long chains of thought. Concurrent work ~\cite{hu2025coordinating,sun2025zerosearch,guan2026deeprag,xiong2025rag,zhao2025r,li-etal-2025-search} combines reasoning with search, inserting external knowledge into long reasoning chains to augment RLMs' reasoning ability. 
These methods integrate external knowledge in a coarse-grained manner, which can introduce instability in the coherence of the reasoning process. Instead, we propose a minimal intervention approach that corrects erroneous knowledge within the model itself, achieving more reliable and controllable knowledge-augmented reasoning capabilities.

\section{Methodology}

\subsection{Preliminary of CheckRLM}
Considering a complicated question $q$ that requires multi-step reasoning, our goal is to generate a correct final answer $a$ by checking and refining the reasoning chain $\mathcal{R}$ generated by RLMs. Mathematically, the reasoning chain $\mathcal{R}$ can be defined as a sequence $\tau = (s_1, s_2, \dots, s_T)$, where each $s_t$ represents an intermediate reasoning state generated at time step $t$. The model parameters $\theta$ govern the conditional distribution $P_\theta(s_t \mid q, s_{<t})$, denoting the probability of generating the subsequent reasoning step $s_t$ conditioned on the question $q$ and the preceding reasoning history $s_{<t}$. Consequently, the sampling process for the entire reasoning chain $\mathcal{R}$ can be formalized as the product of stepwise conditional distributions:
\begin{equation}
    \label{eq:R_conditional_sampling}
    \mathcal{R} \sim \prod_{t=1}^{T} P_\theta(s_t \mid q, s_{<t})
\end{equation}
The final answer $a$ is then derived from both $\mathcal{R}$ and $q$, where $P_\theta(a \mid q, \mathcal{R})$ represents the conditional distribution under which the reasoning model yields the final answer:
\begin{equation}
    \label{eq:a_conditional_sampling}
    a \sim  P_\theta(a \mid q, \mathcal{R})
\end{equation}

\subsection{Knowledge–Thought Coherence Checking}
In this section, we first introduce a two-stage Knowledge–Thought Coherence Checking process (\S\ref{sssec:knowledge_claim_recognition} and \S\ref{sssec:knowledge_coherence_correction}), and then describe the training data construction pipeline and the training process (\S\ref{sssec:optimization}). The overall framework of CheckRLM is shown in Figure~\ref{fig2: Overview of K-COHERE}.

\subsubsection{In-Process Knowledge Claim Recognition}
\label{sssec:knowledge_claim_recognition}
We adopt an ``\textit{intermediate intervention}'' strategy to extract explicit knowledge claims during the reasoning process, rather than intervening only after the complete reasoning chain is generated.
This strategy aims to address potential factual errors and prevent the gradual amplification of randomly occurring erroneous knowledge in the intermediate states of the reasoning chain at an early stage.

Specifically, we identify and extract factual information that supports problem-solving from the reasoning chain, summarizing it into multiple explicit factual claims.
These factual claims serve the fact verification model $\mathcal{M}_{\text{rec}}$ to prioritize question-relevant facts and facilitate the detection of explicit errors for correction, preventing their propagation and negative impact on downstream reasoning.
Considering that granular checks, such as token-level or sentence-level, disrupt the logic and semantic meaning of the reasoning chain and trigger unnecessary or frequent interruptions, we perform coherence checks at the paragraph level, keeping a balance between contextual completeness and timely verification.

Accordingly, the current reasoning chain unit can be defined as the chain generated from $s_{t-1}$ to $s_t$, where the time step $t$ is segmented by paragraphs in the reasoning chain and denoted as $r_t$. We input the original question $q$ and the current reasoning chain unit $r_t$ into the claim recognition model $\mathcal{M}_{\text{rec}}$ to sample a set of knowledge claims $y_{t}^{\text{claim}}$. 
For robustness considerations, we do not use the entire previously generated reasoning chain $\mathcal{R}_{<t}$ as input during recognition to avoid introducing noise from earlier reasoning steps.
We formalize In-Process Knowledge Claim Recognition as follows:
\begin{equation}
    y_{t}^{\text{claim}} \sim \mathcal{M}_{\text{rec}}(\cdot \mid \text{Instruct}_\text{r}, r_t, q)
\end{equation}
In this formulation, $y^{\text{claim}} = \{ y_{t_1}, \ldots, y_{t_n} \}$ represents the set of knowledge claims generated by the claim recognition model $\mathcal{M}_{\text{rec}}$, where $n \in \mathbb{N}^+$ denotes the number of generated claims, and $\text{Instruct}_r$ is a prompt template designed to guide the model to focus on factual claims relevant to the $q$.

\subsubsection{Localized Knowledge Coherence Correction via Retrieval}
\label{sssec:knowledge_coherence_correction}
At the In-Process Knowledge Claim Recognition process, we extract a set of factual claims $y_{t}^{\text{claim}}$ related to the current reasoning chain unit $r_t$, which helps identify specific locations where factual knowledge errors may exist.
Subsequently, the Localized Knowledge Coherence Correction via Retrieval process focuses on correcting the reasoning chain unit $r_t$ at these identified locations by retrieving external knowledge, effectively aligning internal reasoning chain content with external knowledge and correcting parametric knowledge errors.

Specifically, we first take the union of the knowledge claim set $y_{t}^{\text{claim}}$ and the original question $q$ to construct the query set $\mathcal{Q}_t = \{q\} \cup y^{\text{claim}}_{t}$.
Each query \(q_i \in \mathcal{Q}_t\) serves as an atomic retrieval query, triggering an independent retrieval that returns the top-$k$ relevant documents. 
Collecting the results from all queries forms the initial candidate document set \(\mathcal{D}_t^{\text{raw}}\).
To avoid redundancy, we further perform deduplication on the initial candidate document set $\mathcal{D}_t^{\text{raw}}$ to obtain the refined retrieval document set $\mathcal{D}_{t} = \{d_{t_1}, d_{t_2}, \dots, d_{t_n}\}$, where $n \in \mathbb{N}^+$ denotes the number of documents after deduplication.
The process for acquiring external knowledge can be represented in a formal notation as follows:
\begin{equation}
    \mathcal{D}_t = \bigcup_{q_i \in \mathcal{Q}_t} \text{Retriever}(q_i)
\end{equation}

Then, we jointly feed the retrieved document set 
$\mathcal{D}_{t}$ and the current reasoning-chain unit $r_{t}$ into the knowledge correction model $\mathcal{M}_{\text{cor}}$.
The model identifies and extracts knowledge relevant to $r_{t}$ and uses it to assess and verify the factual validity of the current reasoning-chain unit $r_{t}$, ensuring knowledge alignment between the reasoning chain and the knowledge base.
Conditioned on the current reasoning-chain unit \( r_t \) and the retrieved document set \( \mathcal{D}_t \) after deduplication, the corrected reasoning-chain unit \( r_t' \) is sampled from the conditional distribution parameterized by the knowledge correction model \( \mathcal{M}_{\text{cor}} \): 
\begin{equation}
    r_{t}' \sim \mathcal{M}_{\text{cor}}(\cdot \mid \text{Instruct}_\text{c}, r_t, \mathcal{D}_t)
\end{equation}
where $\text{Instruct}_c$ is an instructional prompt template designed to guide the model in performing token-level corrections with minimal cost while ensuring the reasoning structure remains intact.
Here, we introduce two types of preference guidance to steer the model, such that when all facts in $r_t$ are correct or the retrieved document set $\mathcal{D}_t$ is irrelevant, the reasoning-chain unit remains unchanged. If $r_t$ contains erroneous knowledge, the model applies targeted token-level corrections with minimal disruption to preserve the structural integrity of the reasoning chain, ultimately producing the corrected reasoning-chain unit $r_t'$.
We define $\tilde r_t \in \{ r_t, \, r_t' \}$ to uniformly represent the current reasoning chain that admits two possible content states.

After factual errors are corrected, the reasoning model continues the inference process based on the updated reasoning chain, enabling the model to proceed along a correct reasoning path and preventing the accumulation of errors. Through multiple iterations of reasoning and factual correction, the complete reasoning chain is ultimately denoted as:
\begin{equation}
    \mathcal{R} = \tilde r_1 \oplus \tilde r_2 \oplus ... \oplus \tilde r_t
\end{equation}
from which the final answer $a$ is derived. During reasoning, this real-time factual verification and correction mechanism leverages external knowledge to precisely identify and rectify parametric knowledge errors in the RLM, guiding the model toward accurate and efficient reasoning.


\subsubsection{Optimization for Knowledge Recognition and Correction}
\label{sssec:optimization}
Although the two processes above can effectively identify and correct factual errors during reasoning, we also observe that the claim recognition model $\mathcal{M}_{\text{rec}}$ and the knowledge correction model $\mathcal{M}_{\text{cor}}$ suffer from issues in a limited number of cases, such as low-quality factual claims, failure to perform correct corrections, and redundant outputs.
Therefore, our goals are twofold: 1) to improve the factual claim recognition capability of the $\mathcal{M}_{\text{rec}}$, and 2) to enhance the precision of the $\mathcal{M}_{\text{cor}}$ in performing knowledge corrections. 
To this end, we adopt DPO~\cite{rafailov2023direct} as the optimization algorithm to guide the model in assigning higher likelihood to high-quality, accurate, and concise outputs, without requiring an explicit reward model.
Accordingly, we adopt a two-stage strategy to construct a small-scale, high-quality preference dataset $\mathcal{D}_{\text{DPO}}$ (Appendix~\ref{Appendix A: Details of DPO Training Data Construction}), with the aim of inducing preferences for high-quality factual claim recognition and fine-grained knowledge correction.

\noindent\textbf{DPO Training}\quad 
Using the constructed training data, we optimize the model $\mathcal{M}^{\theta}_{\text{RC}}$ that serves both the claim recognition and knowledge correction, to learn a unified policy $\theta$ that assigns higher likelihood to the preferred output $y^{+}$ 
than to its less preferred counterpart $y^{-}$, conditioned on the same input $x$.
Our training objective is formulated as follows:
\begin{multline}
\mathcal{L}_{\text{DPO}}(\mathcal{M}^{\theta}_{\text{RC}};\mathcal{M}^{\text{ref}}_{\text{RC}})=-\mathbb{E}_{\left \{ x,y^{+},y^{-} \right \}\sim \mathcal{D}} [log\sigma \\ [\beta log\frac{\mathcal{M}^{\theta}_{\text{RC}}(y^{+}|x)}{\mathcal{M}^{\text{ref}}_{\text{RC}}(y^{+}|x)}-\beta log\frac{\mathcal{M}^{\theta}_{\text{RC}}(y^{-}|x)}{\mathcal{M}^{\text{ref}}_{\text{RC}}(y^{-}|x)}]]
\label{eq:DPO}
\end{multline}
where $\mathcal{M}^{\theta}_{\text{RC}}$ and $\mathcal{M}^{\text{ref}}_{\text{RC}}$ represent the trainable model and the reference model, respectively.

\section{Experiments}

\subsection{Datasets \& Metrics}
\noindent\textbf{Multi-hop QA Datasets} (1) \textbf{HotpotQA} ~\cite{yang2018hotpotqa} requires retrieving and reasoning over multiple Wikipedia documents with distracting information to answer. (2) \textbf{2WikiMultiHopQA (2WikiMQA)} ~\cite{ho-etal-2020-constructing} is a multi-hop QA dataset where each question requires reasoning over entity-centric hyperlinks across multiple Wikipedia pages to discover bridging entities. (3) \textbf{MuSiQue} ~\cite{trivedi2022musique} is a challenging dataset characterized by its compositional, interconnected questions that necessitate sequential reasoning without relying on isolated single-hop clues. (4) \textbf{IIRC} ~\cite{ferguson-etal-2020-iirc} builds on Wikipedia articles where questions require identifying relevant text snippets from linked contexts to answer questions given an incomplete context. For the test samples and retrieval corpus, we follow IRCoT ~\cite{trivedi2023interleaving}.\\
\noindent\textbf{Short-form QA Dataset}\quad focuses on retrieving direct answers to factoid questions from a single context. We choose \textbf{SimpleQA} ~\cite{wei2024measuring}, a large-scale benchmark consisting of short, fact-seeking questions. We randomly select 500 samples from the test split and use the KILT ~\cite{petroni2021kilt} knowledge source as the corpus. \\
\noindent\textbf{Metrics} \quad We use F1-Score (f1) and Exact Match (em) ~\cite{jiang2023active} to evaluate the correctness of the final answer.

\subsection{Baselines}
We compare our method against the following baseline approaches: (1) Direct Reasoning: the RLM relies solely on its parametric knowledge to answer questions without retrieving any external information.
(2) Vanilla RAG: a single-step retrieval fetches the top-$k$ most relevant documents, which are then provided alongside the question as context for the LLM to generate an answer.
(3) Multi-Step RAG: the RAG system gathers relevant information through multiple iterative retrieval steps. We select RAT ~\cite{li2024rat} as our representative baseline.
(4) Adaptive RAG: the RAG framework dynamically decides the retrieval and generation strategy to optimize both response quality and efficiency. We primarily compare against three established baselines: FLARE ~\cite{jiang2023active}, Self-RAG ~\cite{asai2024self}, and ReAct ~\cite{yao2022react}. Additionally, we also compare with Search-o1 ~\cite{li-etal-2025-search}, which actively retrieves external knowledge during the reasoning process and seamlessly integrates refined, concise information extracted from lengthy retrieval results into the reasoning chain.

\begin{table*}[ht]
  \centering
    \begin{tabular}{lccccccccccc}
    \toprule
    \multirow{3}[3]{*}{\textbf{Method}} & \multicolumn{8}{c}{\textbf{Multi-hop}} &  \multicolumn{2}{c}{\textbf{Short-form}} & \multicolumn{1}{c}{} \\
    \cmidrule(lr){2-9}\cmidrule(lr){10-11}
    & \multicolumn{2}{c}{\textbf{HotpotQA}} & \multicolumn{2}{c}{\textbf{2WikiMQA}} & \multicolumn{2}{c}{\textbf{MuSiQue}} & \multicolumn{2}{c}{\textbf{IIRC}} & \multicolumn{2}{c}{\textbf{SimpleQA}} & \textbf{Avg.} \\
    \cmidrule(lr){2-3}\cmidrule(lr){4-5}\cmidrule(lr){6-7}\cmidrule(lr){8-9}\cmidrule(lr){10-11}\cmidrule(lr){11-11}
    & f1 & em & f1 & em & f1 & em & f1 & em & f1 & em & \\
    \midrule
    \multicolumn{12}{l}{\textit{QwQ-32B}} \\
    \hdashline
    Direct Reasoning & 38.4 & 29.0 & 34.6 & 28.4 & 18.5 & 7.4 & 24.8 & 21.4 & 10.5 & 4.6 & 21.8 \\
    Vanilla RAG & 52.7 & 42.2 & 46.4 & 42.6 & 19.3 & 10.0 & 25.0 & 22.0 & 31.4 & 24.2 & 31.6\\
    ReAct (\citeyear{yao2022react}) & 48.2 & 34.6 & 45.9 & 33.0 & 22.3 & 10.0 & 21.9 & 16.2 & 30.1 & 22.2 & 28.4 \\
    FLARE (\citeyear{jiang2023active}) & 43.5 & 32.2 & 46.6 & 38.4 & 18.4 & 9.2 & 12.3 & 9.6 & 24.1 & 17.4 & 25.2\\
    Self-RAG (\citeyear{asai2024self}) & 49.7 & 39.9 & 40.0 & 36.2 & 20.0 & 10.6 & 25.1 & 21.3 & 31.1 & 24.8 & 29.9\\
    RAT (\citeyear{li2024rat}) & 51.5 & 39.0 & 47.2 & 38.6 & 24.1 & 13.2 & 21.7 & 17.0 & 30.3 & 22.2 & 30.5\\
    Search-o1 (\citeyear{li-etal-2025-search}) & 62.0 & 49.2 & 71.4 & 60.4 & 33.3 & 20.7 & 29.2 & 25.0 & 35.4 & 27.4 & 41.5 \\
    CheckRLM & \textbf{66.3} & \textbf{52.6} & \textbf{73.4} & \textbf{62.0} & \textbf{39.6} & \textbf{27.2} & \textbf{33.1} & \textbf{29.0} & \textbf{40.0} & \textbf{30.4} & \textbf{45.4}\\
    \midrule
    \multicolumn{12}{l}{\textit{Qwen3-32B}} \\
    \hdashline
    Direct Reasoning & 36.4 & 27.6 & 34.0 & 29.4 & 16.6 & 6.6 & 23.0 & 20.2 & 9.5 & 2.4 & 20.6\\
    Vanilla RAG & 51.2 & 41.0 & 44.5 & 41.8 & 19.7 & 10.4 & 22.5 & 19.8 & 29.3 & 22.8 & 30.3\\
    CheckRLM & \textbf{64.2} & \textbf{52.0} & \textbf{70.8} & \textbf{60.8} & \textbf{35.6} & \textbf{25.4} & \textbf{31.6} & \textbf{27.8} & \textbf{39.1} & \textbf{30.6} & \textbf{43.8}\\
    \midrule
    \multicolumn{12}{l}{\textit{Qwen3-8B}} \\
    \hdashline
    Direct Reasoning & 29.3 & 22.2 & 31.1 & 26.6 & 12.9 & 5.2 & 21.0 & 18.2 & 7.7 & 2.8 & 17.7\\
    Vanilla RAG & 44.5 & 35.2 & 40.9 & 38.4 & 13.8 & 6.0 & 19.8 & 16.8 & 29.4 & 23.0 & 26.8\\
    CheckRLM & \textbf{61.9} & \textbf{49.6} & \textbf{69.7} & \textbf{58.6} & \textbf{35.0} & \textbf{24.0} & \textbf{29.7} & \textbf{25.4} & \textbf{35.7} & \textbf{28.6} & \textbf{41.8}\\
    
    \bottomrule
    \end{tabular}%
  \caption{\textbf{Results (\%) of overall performance.} Our recognition and correction model is Llama-3.3-70B-Instruct. The highest scores are emphasized in \textbf{bold}.}
  \label{tab: main results}%
\end{table*}%

\subsection{Implementation Details}
We primarily employ QwQ-32B ~\cite{qwq32b} as the reasoning model, while also evaluating various RLMs of Qwen3-8B ~\cite{yang2025qwen3}, Qwen3-32B ~\cite{yang2025qwen3} and DeepSeek-R1-Distill-Llama-70B ~\cite{guo_deepseek-r1_2025} for comparative analysis.
For $\mathcal{M}_{\text{rec}}$ and $\mathcal{M}_{\text{cor}}$, we use the same underlying model and select four LLMs: Qwen3-8B ~\cite{yang2025qwen3}, Qwen2.5-14B-Instruct ~\cite{qwen2.5}, Qwen2.5-32B-Instruct ~\cite{qwen2.5}, and Llama-3.3-70B-Instruct ~\cite{dubey2024llama}. In terms of inference hyperparameters, we set the maximum context length to 16,384 tokens and retain default settings for all other hyperparameters. For the retriever, we utilize both the sparse retriever BM25 and the dense retriever bge-large-en-v1.5 ~\cite{bge_embedding} on multi-hop QA datasets, while only the dense retriever bge-large-en-v1.5 is used on the short-form QA dataset. Across all experiments, we fix top-$k$ to 3 and set the maximum number of recognition-and-correction steps to 10. All experiments are conducted using the vLLM ~\cite{kwon2023efficient} on a cluster of 8×A800 GPUs. For DPO training, we use a batch size of 8, a learning rate of 5e-7, $\beta$ of 0.1, and train for 1 epoch.

\section{Results and Analysis}
\subsection{Overall Performance}

\textbf{Vanilla RAG underperforms on complex tasks; by contrast, CheckRLM demonstrates a substantial performance improvement.}
As shown in Table~\ref{tab: main results}, Vanilla RAG improves upon Direct Reasoning by introducing external knowledge documents in a one-time retrieval step, thus partially alleviating the limitation that Direct Reasoning relies entirely on internal knowledge during the reasoning process. Empirical results indicate that Vanilla RAG achieves reasonable performance on short-form QA tasks, as these tasks typically require only a small number of relevant snippets that are most relevant to the original question; however, these observed performance improvements struggle to generalize to complex multi-hop QA tasks.
In contrast, CheckRLM consistently outperforms Vanilla RAG across all datasets. In particular, it achieves especially substantial gains on complex multi-hop QA tasks, yielding improvements over Vanilla RAG across all three reasoning backbones. These performance improvements indicate that CheckRLM is capable of performing effective knowledge correction by repeatedly and timely incorporating external knowledge to precisely correct factual errors in the reasoning chain. 

\textbf{CheckRLM surpasses a wide range of mainstream baselines across both multi-hop and short-form QA benchmarks.}
\begin{table}[ht]
  \centering
  \renewcommand{\arraystretch}{0.9}
  \begin{tabular}{lccc}
    \toprule
    \textbf{Method} & \textbf{Hot.} & \textbf{2Wiki.} & \textbf{Simp.} \\
    \midrule
    
    \multicolumn{4}{l}{\textit{QwQ-32B}} \\
    \hdashline
    Vanilla RAG     & 52.7 & 46.4 & 31.4 \\
    Inference Only  & 61.6 & 65.5 & 36.0 \\
    DPO Training    & \textbf{63.2} & \textbf{71.2} & \textbf{36.9} \\
    
    \midrule
    
    \multicolumn{4}{l}{\textit{Qwen3-32B}} \\
    \hdashline
    Vanilla RAG     & 51.2 & 44.5 & 29.3 \\
    Inference Only  & 57.0 & 62.9 & 34.2 \\
    DPO Training    & \textbf{61.3} & \textbf{65.3} & \textbf{36.1} \\
    \bottomrule
  \end{tabular}
  \caption{\textbf{DPO Training Results (\%).} We conduct experiments on the HotpotQA, 2WikiMQA and SimpleQA dataset, using Qwen2.5-14B-Instruct as the recognition and correction model and f1 as the metric.}
  \label{tab:dpo_results_f1_small}
\end{table}
As shown in Table~\ref{tab: main results}, in comparison with recent iterative retrieval–generation and coupled retrieval–reasoning methods, CheckRLM achieves the highest scores across all evaluated datasets, including challenging multi-hop and short-form QA benchmarks. The gains are especially pronounced on complex multi-hop reasoning, and are most evident on the more challenging MuSiQue dataset, where CheckRLM outperforms the best-performing baseline by 6.3 f1 points.
This advantage can be attributed to a key design difference.
While these prior methods enhance reasoning by expanding the interaction between retrieval and reasoning or by increasing search depth, they generally lack explicit mechanisms to correct factual inconsistencies within the reasoning process itself.
In contrast, CheckRLM introduces an in-process knowledge checking mechanism that continuously monitors and corrects factual errors during reasoning, preventing erroneous knowledge from propagating across subsequent reasoning steps. This design is especially critical for multi-hop reasoning, where early-stage errors can severely affect downstream inference. 

\subsection{Effectiveness of Knowledge Recognition and Correction Optimization}
\label{ssec:DPO Training Results}
The DPO training process in CheckRLM enhances the model’s ability to correct factual errors in the reasoning chain by jointly optimizing knowledge claim recognition and knowledge coherence correction. Based on this design, we evaluate the performance gains brought by this joint optimization strategy to validate the effectiveness of our training procedure.
We conduct DPO training on the small backbone Qwen2.5-14B-Instruct, and the results are summarized in Table~\ref{tab:dpo_results_f1_small}. Compared to inference-only reasoning, DPO yields a 5.7\% improvement on the in-domain dataset 2WikiMQA, while improving performance on out-of-domain multi-hop QA benchmarks such as HotpotQA, as well as the short-form QA dataset SimpleQA.

Furthermore, we perform extensive experiments across different reasoning backbones and training data compositions, where CheckRLM also achieves strong performance on all evaluated datasets (see Appendix~\ref{ssec:Knowledge Recognition and Correction Optimization Results} for complete results). These results demonstrate that our DPO data sampling and training pipeline effectively enhances CheckRLM across diverse reasoning models and datasets, exhibiting strong generality and robustness.

\begin{figure}[ht]
    \centering
    \begin{subfigure}{0.49\columnwidth}
        \centering
    \includegraphics[width=\linewidth]{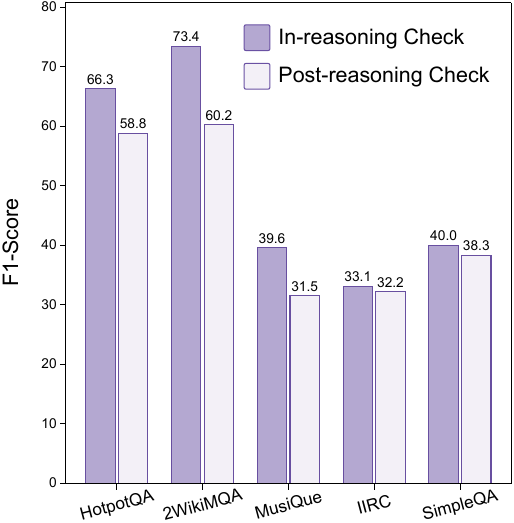}
        \caption{Llama-3.3-70B-Instruct}
        \label{fig_a:on-offline-70B}
    \end{subfigure}%
    \hfill 
    \begin{subfigure}{0.49\columnwidth}
        \centering
    \includegraphics[width=\linewidth]{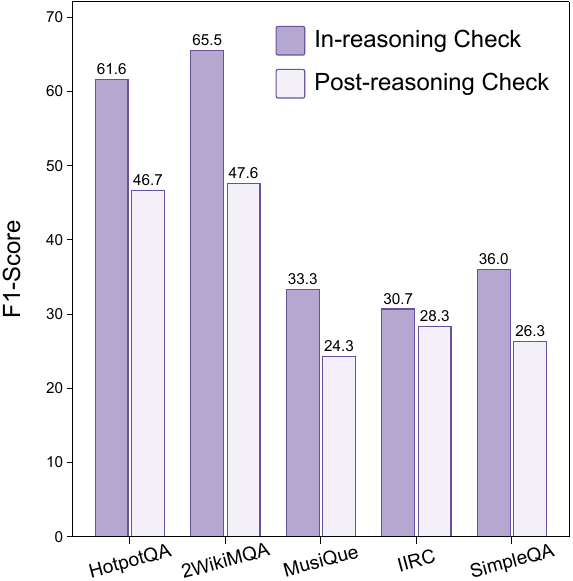}
        \caption{Qwen2.5-14B-Instruct}
        \label{fig_b:on-offline-14B}
    \end{subfigure}

    \caption{\textbf{Post-reasoning Check vs. In-reasoning Check.} We use QwQ-32B as the reasoning model, Llama-3.3-70B-Instruct and Qwen2.5-14B-Instruct as the recognition and correction model. The experiments are conducted in the 2WikiMQA dataset and use f1 as the metric. More results are in Appendix~\ref{ssec: Post‑reasoning Check vs. In‑reasoning Check Results}. }
    \label{fig:online-offline-f1-ab}
\end{figure}
\subsection{Effectiveness of In-reasoning Knowledge Check in CheckRLM}
CheckRLM introduces an intermediate intervention strategy that integrates knowledge claim recognition with consistency correction for in-reasoning check. Benchmarked against post-reasoning check, CheckRLM consistently outperforms across all datasets (Figure~\ref{fig:online-offline-f1-ab}).
The results confirm the superiority of in-reasoning intervention. 

RLMs often introduce errors early in inference, which then propagate and amplify through subsequent steps. While post-reasoning check accesses the complete chain, its late correction can only address a subset of errors and fails to rectify already flawed reasoning trajectories.
In contrast, CheckRLM intervenes dynamically during reasoning—correcting factual errors before they cascade and ensuring that each step proceeds under accurate knowledge constraints. This step-wise correction substantially improves the stability and reliability of the overall reasoning process.

\begin{figure}[htbp]
    \centering
  \includegraphics[width=0.9\columnwidth]{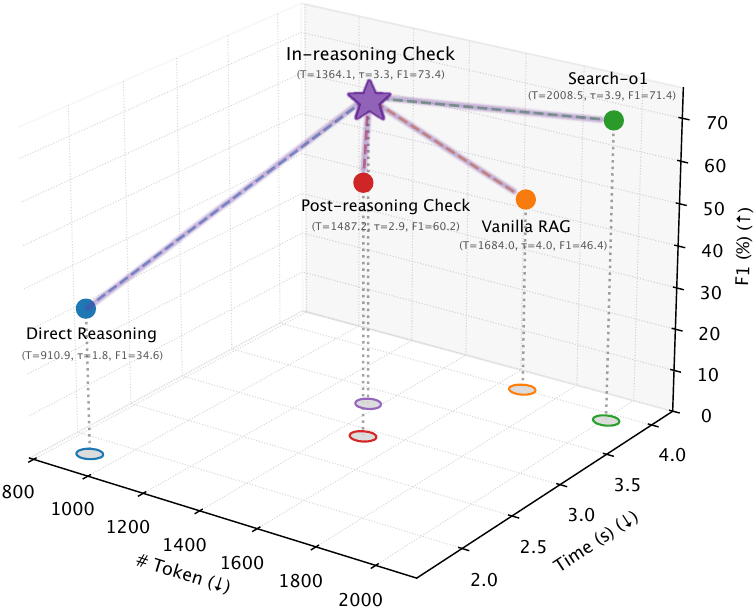}
  \caption{\textbf{Time, token consumption and performance of different methods.} We use QwQ-32B as the reasoning model, Llama-3.3-70B-Instruct as the recognition and correction model. The experiments are conducted in the 2WikiMQA dataset. More results are in Appendix~\ref{ssec:Analysis of Cost-Efficient Reasoning Results}.}
  \label{fig:Cost 3D}
\end{figure}
\subsection{Analysis of Cost-Efficient Reasoning}
In this experiment, we conduct a three-dimensional analysis that jointly considers token consumption, inference time, and task performance, aiming to provide a comprehensive evaluation of efficiency and performance.

As shown in Figure~\ref{fig:Cost 3D}, Direct Reasoning has the lowest cost but limited performance. In contrast, Vanilla RAG and Search-o1 improve performance at substantially higher token and time costs.
Notably, the in-reasoning check in CheckRLM attains a more favorable outcome across both performance and cost dimensions. Compared with Vanilla RAG and Search-o1, it significantly reduces token consumption and inference time while simultaneously achieving the best performance. This suggests that timely knowledge correction during reasoning enables the model to avoid unnecessary reasoning steps and redundant retrieval caused by error accumulation, improving efficiency without sacrificing accuracy.

\begin{figure}[ht]
    \centering
    \begin{subfigure}{0.49\columnwidth}
        \centering
    \includegraphics[width=\linewidth]{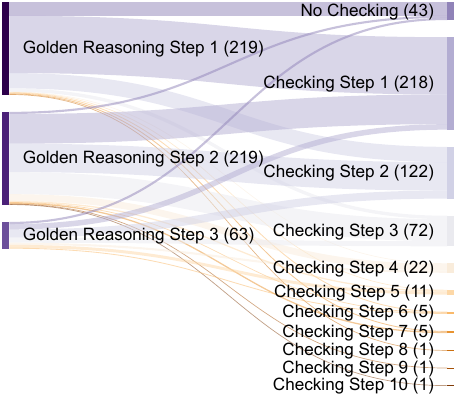}
        \caption{2WikiMQA}
        \label{fig_a:check-refine-70B}
    \end{subfigure}%
    \hfill 
    \begin{subfigure}{0.49\columnwidth}
        \centering
    \includegraphics[width=\linewidth]{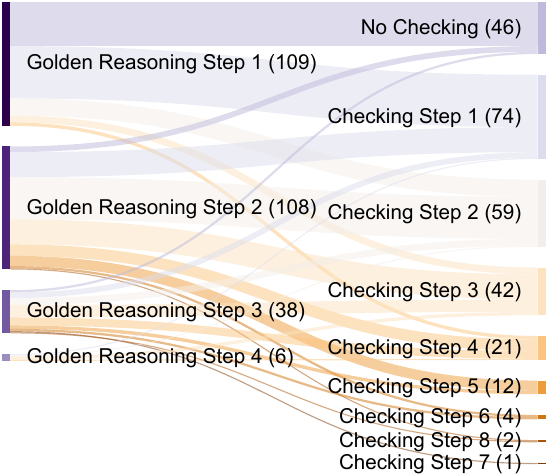}
        \caption{MuSiQue}
        \label{fig:correction_distribution_analysis-70B}
    \end{subfigure}
    \caption{\textbf{Correspondence between Golden Reasoning Step and Checking Step.} We use QwQ-32B as the reasoning model, Llama-3.3-70B-Instruct as the recognition and correction model. The experiments are conducted in the 2WikiMQA and MuSiQue datasets.}
    \label{fig:correction_distribution_analysis}
\end{figure}
Furthermore, the comparison between the in-reasoning check in CheckRLM and the post-reasoning check highlights the critical role of correction timing. Although post-reasoning check reduces computational cost relative to retrieval-heavy methods, it remains less effective than in-process check, as delayed correction cannot prevent erroneous knowledge from propagating throughout the reasoning chain.
Overall, CheckRLM achieves the best performance while substantially reducing both inference time and token consumption, demonstrating clear superiority over all competing methods in terms of both efficiency and effectiveness.

\subsection{Characteristics of Correction Behavior}
This experiment aims to analyze when factual errors in multi-hop reasoning are detected and corrected by our method using the GPT-4o model, relative to the golden reasoning steps.

As depicted in Figure~\ref{fig:correction_distribution_analysis}, Golden Reasoning Step $k$ denotes the $k$-th hop in the golden reasoning chain, while Checking Step $j$ indicates that our method successfully identifies and corrects the error at its $j$-th checking iteration.
The observed trend shows that corrections predominantly occur at early checking steps (Steps 1–2), indicating effective early interception of factual errors rather than their propagation through the reasoning chain.

Despite its higher reasoning complexity, MuSiQue still exhibits a clear tendency toward early correction. While Checking Step 1 remains dominant, errors arising from later oracle steps are primarily resolved at intermediate stages (Steps 2–4), with deep checking being rare. This indicates that the method consistently corrects factual errors at early stages, even under more challenging multi-hop reasoning tasks. These results demonstrate that our method performs early, targeted, and efficient error correction, rather than relying on deep search or repeated post-hoc refinement.

\section{Conclusion}
In conclusion, we present \textbf{CheckRLM}, a framework that tackles factual errors in RLMs for knowledge-intensive tasks. CheckRLM enhances reasoning reliability by proactively recognizing and correcting factual inconsistencies during inference via RAG. It extracts factual claims, locates errors against external knowledge, and performs precise, low-cost corrections. Experiments demonstrate that CheckRLM significantly outperforms existing baselines, mitigates error accumulation in long-horizon reasoning, and reduces temporal and computational costs, advancing towards more accurate and efficient knowledge-reliant reasoning.

\section*{Limitations}
Although CheckRLM has successfully achieved real-time intervention of external knowledge for reasoning verification, this study still has certain limitations, mainly reflected in two aspects. First, the current mechanism focuses on utilizing pure textual knowledge sources and has not yet been extended to the collaborative verification of multimodal information. Second, the framework is currently optimized only for a single knowledge base and does not explore the complex coordination and conflict resolution mechanisms across multiple heterogeneous knowledge bases. Future research could delve into the directions of multimodal fusion and dynamic integration of multi-source knowledge to build a more general and robust reasoning verification framework.

\section*{Acknowledgments}
This work was supported by the National Natural Science
Foundation of China (No. 62276026) and the Fundamental Research Funds for the Central Universities (No. 2253500001). This work is supported by the AI9Stars community.

\bibliography{acl}

\clearpage

\appendix
\section*{Appendix}
\label{sec:appendix}

\startcontents[sections]
\printcontents[sections]{l}{1}{\setcounter{tocdepth}{2}}

\section{Details of DPO Training Data Construction}
\label{Appendix A: Details of DPO Training Data Construction}
We randomly select 2500 samples from the 2WikiMQA training set to build our DPO training dataset $\mathcal{D}_{\text{DPO}}$, which comprises two subsets: training data for knowledge claim recognition $\mathcal{D}_{\rm \text{KCR}}$ and training data for knowledge coherence correction $\mathcal{D}_{\rm \text{KCC}}$. The statistics of all DPO training data are presented in Table~\ref{tab:statistics_DPO}.

\subsection{DPO Training Data for Knowledge Claim Recognition}
\label{ssec:Knowledge Claim Recognition DPO training data}
Observing that knowledge corrections primarily occur in the initial steps of the reasoning process, we only sample data from the first three reasoning chain units to efficiently obtain high-quality data.
For a given sample, we input the original question $q$, the current reasoning chain unit $r_t$, and the recognition instruction template $\text{Instruct}_r$ into $\mathcal{M}_{\text{rec}}$ to sample a set of knowledge claims $y_t^{\text{claim}}$.
To balance diversity and accuracy in the generated samples, we set different decoding hyperparameters, namely temperature and $\text{top}\_p$, to specific values of 0.1, 0.5, and 0.9. They control the generation of a candidate answer set $\mathcal{Y}^{\text{claim}}_{t} = \{y^{\text{claim}}_{t_1}, y^{\text{claim}}_{t_2}, ..., y^{\text{claim}}_{t_n}\}$, where $n\in \mathbb{N}^+$ denotes the number of candidate answers. 
Recent studies have demonstrated that annotations from advanced models can achieve high alignment with human judgments ~\cite{liu-etal-2023-g,fu-etal-2024-gptscore}. We utilize GPT-4o-mini to annotate positive and negative examples $y_{t}^{\text{claim(+)}}$ and $y_{t}^{\text{claim(-)}}$ within the candidate answer set. The resulting training data $\mathcal{D}_{\rm \text{KCR}}$ can be represented as $\{x_t, y_{t}^{\text{claim(+)}}, y_{t}^{\text{claim(-)}}\}$. 

\begin{table}[t]
\centering
\begin{tabular}{l|lll} 
\hline
  & \(\mathcal{D}_{\rm \text{KCR}}\) & \(\mathcal{D}_{\rm \text{KCC}}\) &  \(\mathcal{D}_{\text{DPO}}\) \\ 
\hline
\textbf{\# Sample} & 2351 & 2960 & 5311 \\ 
\hline
\end{tabular}
\caption{Statistics of DPO Training Data.}
\label{tab:statistics_DPO} 
\end{table}
\subsection{DPO Training Data for Knowledge Coherence Correction}
\label{ssec:Knowledge Coherence Correction DPO training data}
For training data $\mathcal{D}_{\rm \text{KCC}}$, the settings for decoding hyperparameters remain the same as above. Since the quality of retrieved documents is positively correlated with the quality of knowledge claims, we use the positive example $y_{t}^{\text{claim(+)}}$ for document retrieval.
The retrieved document set $\mathcal{D}_t^+$, the current reasoning chain unit $r_t$, and the correction instruction template $\text{Instruct}_\text{c}$ are then fed into $\mathcal{M}_{\text{cor}}$ to generate a candidate answer set $\tilde{\mathcal{R}_t} = \{\tilde r_{t_1}, \tilde r_{t_2}, ..., \tilde r_{t_n}\}$, where $n\in \mathbb{N}^+$ is the number of candidate answers.
Similarly, we employ GPT-4o-mini to annotate positive and negative examples $\tilde r_t^{+}$ and $\tilde r_t^{-}$ and construct the DPO training data for knowledge
coherence correction, denoted as $\{x_t, \tilde r_t^{+}, \tilde r_t^{-}\}$.
We then use the positive example $\tilde r_t^{+}$ to form the reasoning chain $\mathcal{R}$ of the RLM to generate the next reasoning chain unit $r_{t+1}$.

\section{Prompt Details}
\label{Appendix B: Prompt Details}
This section presents the prompt templates employed for both inference and DPO training. For inference, we provide prompts for Direct Reasoning, Vanilla RAG, knowledge claim recognition, knowledge coherence correction and correction behavior analysis. 
In particular, the reasoning prompt used in CheckRLM is identical to the direct reasoning. For DPO training, we supply corresponding prompts for evaluating knowledge claim recognition and knowledge coherence correction.

\subsection{Direct Reasoning Prompt}
\begin{tcolorbox}[
    colframe=black,       
    colback=blue!10!white,             
    coltitle=white,                   
    coltext=black,                    
    fonttitle=\bfseries,              
    title=Direct Reasoning Prompt,  
    boxrule=1pt,                      
    arc=2mm,                          
    width=\linewidth,                 
    left=7pt,                         
    right=7pt,                        
    top=5pt,                          
    bottom=5pt                        
]
\fontsize{8.5pt}{10pt}\selectfont
Please reason step by step, and put your final answer within \textbackslash{}boxed\{\}. Only give me the answer and do not output any other words.

Question:  \{question\}
\end{tcolorbox}

\subsection{Vanilla RAG Prompt}
\begin{tcolorbox}[
    colframe=black,       
    colback=blue!10!white,             
    coltitle=white,                   
    coltext=black,                    
    fonttitle=\bfseries,              
    title=Vanilla RAG Prompt,  
    boxrule=1pt,                      
    arc=2mm,                          
    width=\linewidth,                 
    left=7pt,                         
    right=7pt,                        
    top=5pt,                          
    bottom=5pt                        
]
\fontsize{8.5pt}{10pt}\selectfont
Answer the question based on the given passages. Please reason step by step, and put your final answer within \textbackslash{}boxed\{\}. Only give me the answer and do not output any other words.

The following are given passages:

\{refs\}

Question:  \{question\}
\end{tcolorbox}

\subsection{CheckRLM Prompt}

\begin{tcolorbox}[
    colframe=black,       
    colback=blue!10!white,             
    coltitle=white,                   
    coltext=black,                    
    fonttitle=\bfseries,              
    title=Knowledge Claim Recognition Prompt,  
    boxrule=1pt,                      
    arc=2mm,                          
    width=\linewidth,                 
    left=7pt,                         
    right=7pt,                        
    top=5pt,                          
    bottom=5pt                        
]
\fontsize{8.5pt}{10pt}\selectfont
Task: Extract only explicit factual claims from the Reasoning Process that are helpful to answering the Question. The Reasoning Process contains factual information useful for addressing the Question. Your task is to identify, extract, and summarize these relevant facts into clear factual claims. The factual claims must be based on the Reasoning Process; NEVER add any new content by yourself. Try to avoid two factual claims that have similar meaning. Output is a list of strings. The output format is as follows: 

["factual claim 1", "factual claim 2", ...]

Question:  \{question\}

Reasoning Process: \{reasoning\}

Factual Claim List:
\end{tcolorbox}

\begin{tcolorbox}[
    colframe=black,       
    colback=blue!10!white,             
    coltitle=white,                   
    coltext=black,                    
    fonttitle=\bfseries,              
    title=Knowledge Coherence Correction Prompt,  
    boxrule=1pt,                      
    arc=2mm,                          
    width=\linewidth,                 
    left=7pt,                         
    right=7pt,                        
    top=5pt,                          
    bottom=5pt                        
]
\fontsize{8.5pt}{10pt}\selectfont
Task: Based on the retrieved documents, check and correct factual errors in the Reasoning Process. You will receive several retrieved documents and the Reasoning Process. The retrieved documents may contain information relevant to the Reasoning Process. Your task is to filter the relevant information from the retrieved documents and verify the correctness of the factual information in the Reasoning Process. Follow these rules:

1. If the factual information in the Reasoning Process is correct, only output the original Reasoning Process.

2.  If the factual information in the Reasoning Process is incorrect, make the minimal necessary corrections to fix the error without altering the structure or flow of the Reasoning Process.

3. If the retrieved documents do not contain any relevant information, only output the original Reasoning Process.

NEVER add any supplementary information from the retrieved documents. Only correct factual errors when necessary. Only give me your modified reasoning process and do not output any other words.

Retrieved documents: \{refs\}

Reasoning Process: \{reasoning\}

Provide your modified reasoning process:
\end{tcolorbox}

\subsection{Correction Behavior Prompt}
\begin{tcolorbox}[
    colframe=black,
    colback=blue!10!white,
    coltitle=white,
    coltext=black,
    fonttitle=\bfseries,
    title=Correction Behavior Prompt,
    boxrule=1pt,
    arc=2mm,
    width=\linewidth,
    left=7pt,
    right=7pt,
    top=5pt,
    bottom=5pt
]
\fontsize{8.5pt}{10pt}\selectfont
Task: Label Effective Correction Steps

You are a data annotation expert tasked with identifying which refinement steps successfully introduce key content from the correct reasoning chain.

Requirements:

1. Examine the ``original think'' and ``refine think'' content of each step one by one. 

2. If ``original think'' content contains the key step from the correct reasoning chain, output the correct reasoning step number and -1.

3. If the ``refine think'' content contains any key step from the correct reasoning chain, output the correct reasoning step number and the step number of the modified content.

4. If multiple steps contain the same key step, only record the earliest occurrence.

Output Rules:

1. Output format: [\{``Correct Reasoning Step Number'': ``First Modefied Number''\}]

2. Do not include any explanations or additional text. Only output the list.

Input:

Question: \{query\}

Correct Reasoning Chains: \{golden\_reasoning\_steps\}

Modification Records: \{think\_refine\_detail\}
\end{tcolorbox}

\subsection{DPO Evaluation Prompt}
\begin{tcolorbox}[
    colframe=black,
    colback=blue!10!white,
    coltitle=white,
    coltext=black,
    fonttitle=\bfseries,
    title=Knowledge Claim Recognition Evaluation Prompt,
    boxrule=1pt,
    arc=2mm,
    width=\linewidth,
    left=7pt,
    right=7pt,
    top=5pt,
    bottom=5pt
]
\fontsize{8.5pt}{10pt}\selectfont
Task: Evaluate factual claim lists generated based on a Reasoning Process and Question. Your task is to select the best factual claim list and the worst factual claim list according to Quality criteria strictly. Quality criteria refer to:

1. Relevance: each factual claim in the factual claim list is relevant to the Question and Reasoning Process, note that if there is no factual claim in Reasoning Process, just output [] but not similar sentences such as \textquotedblleft there is no related information\textquotedblright\ or \textquotedblleft the reasoning process does not contain any factual claim\textquotedblright;

2. Specificity: each factual claim in the factual claim list is clear, avoiding unclear pronoun;

3. No redundancy: The correct factual claim list format is just one list of strings, no multiple lists or explanatory Notes.

Output Rules:

1. JSON format: \{\textquotedblleft best\_id\textquotedblright{}: <id of the factual claim list that fully meets all Quality criteria above>, \textquotedblleft worst\_id\textquotedblright{}: <id of the factual claim list that violates the most Quality criteria>\};

2. If there is a minimal quality difference between the best factual claim list and the worst factual claim list or there is no worst factual claim, assign the same id to both best\_id and worst\_id;

3. If no factual claim list meets all the Quality criteria above, return an empty object \{\}

4. Do not include any explanations or additional text.

Question: \{question\}

Reasoning Process: \{reasoning\}

Generated factual claim lists: \{check\_responses\}
\end{tcolorbox}

\begin{tcolorbox}[
    colframe=black,
    colback=blue!10!white,             
    coltitle=white,
    coltext=black,
    fonttitle=\bfseries,
    title=Knowledge Coherence Correction Evaluation Prompt,  
    boxrule=1pt,
    arc=2mm,
    width=\linewidth,
    left=7pt,
    right=7pt,
    top=5pt,
    bottom=5pt
]
\fontsize{8.5pt}{10pt}\selectfont
Task: Evaluate refined reasoning processes based on quality criteria strictly and identify the best and the worst refined reasoning processes. Quality criteria refer to:

1. Structural Integrity: strictly preserves the original reasoning process structure. Do not add any other supplement information at the end of the original reasoning process, even if the supplement information is right. The refined reasoning process contains negative sentences such as \textquotedblleft the retrieved documents do not contain related information\textquotedblright\ is the worst;

2. Precise Corrections: Only modifies factually incorrect content verified by retrieved documents.

3. Conciseness: No redundant text in the refined reasoning process, including unnecessary prefixes, explanatory Notes, and continuation after the original reasoning process, such as \textquotedblleft According to the retrieved documents\textquotedblright.

Output Rules:

1. JSON format: \{\textquotedblleft best\_id\textquotedblright{}: <id of the refined reasoning process that fully meets all Evaluation Criteria>, \textquotedblleft worst\_id\textquotedblright{}: <id of the refined reasoning process that violates the most Quality criteria>\}

2. If there is a minimal quality difference between the best reasoning process and the worst reasoning process or there is no worst refined reasoning process, assign the same id to both best\_id and worst\_id

3. If no refined reasoning process meets all Evaluation Criteria above, return an empty object \{\}

4. Do not include any explanations or additional text.

Retrieved documents: \{refs\}

Original reasoning process: \{think\}

Generated refined reasoning processes: \{refine\_responses\}
\end{tcolorbox}

\section{Additional Experimental Results}
\label{Appendix C: Additional Experimental Results}
\subsection{Overall Performance of CheckRLM}
\label{ssec:Overall Performance of CheckRLM}

The Table~\ref{tab: appendix main results} presents overall results of various RLMs and recognition and correction models. By combining reasoning models and recognition and correction models of varying parameter sizes, we find that CheckRLM consistently achieves better performance than both Direct Reasoning and Vanilla RAG across all datasets, demonstrating the generality and robustness of CheckRLM.

\subsection{Knowledge Recognition and Correction Optimization}
\label{ssec:Knowledge Recognition and Correction Optimization Results}
The Table~\ref{tab-app: dpo results} presents all results of CheckRLM with DPO training. Using the Qwen2.5-14B-Instruct model as the backbone, we conduct experiments across different reasoning models and datasets. The results show that DPO training outperforms Inference Only in almost all cases, demonstrating the effectiveness and stability of our constructed training data pipeline and training.

To further analyze the impact of different data compositions, we conduct additional experiments (Table~\ref{tab-app: different dpo results}) and find that DPO training on the combined data $\mathcal{D}_{\rm \text{KCR}}$+$\mathcal{D}_{\rm \text{KCC}}$ outperforms training on single-type data, demonstrating the effectiveness of both $\mathcal{D}_{\rm \text{KCR}}$ and $\mathcal{D}_{\rm \text{KCC}}$. 
Furthermore, the performance improvement brought by $\mathcal{D}_{\rm \text{KCR}}$ is more pronounced in most cases, which can be attributed to the fact that higher-quality factual claims enable the correction model to rectify factual errors more accurately.

\subsection{Constrained Ablation Study}
\label{ssec: Ablation Study}
In the constrained ablation study, given the high interdependence between the In-Process Knowledge Claim
Recognition and Localized Knowledge Coherence Correction modules that prevent their complete removal, we apply the simplest versions of both modules and conduct experiments across various models. The experimental results, as shown in the Table~\ref{tab: ablation}, demonstrate the importance of both modules within the CheckRLM framework.

\subsection{Post‑reasoning Check vs. In‑reasoning Check}
\label{ssec: Post‑reasoning Check vs. In‑reasoning Check Results}
The Table~\ref{tab-app: local_global} compares the complete results of Post‑reasoning Check and In‑reasoning Check under various combinations of RLMs and recognition and correction models. The In‑reasoning Check results outperform the Post‑reasoning Check method across all experimental settings, demonstrating the superiority of CheckRLM.

\subsection{Effect of the Retrieved Summary Injection on Reasoning}
\label{ssec: Summary Results}

The goal of CheckRLM correction is to correct errors in the reasoning chain with minimal cost, while preserving the original distribution of the reasoning model as much as possible.
Our observations indicate that the reasoning model itself exhibits a clear and coherent problem-solving approach, and that incorporating additional information may interfere with its inherent reasoning logic.
To investigate this phenomenon, we design a comparative experiment that injects summarized retrieved information into the reasoning chain during error correction.

The experimental results are presented in the Table~\ref{tab-app: Summary}. We observe that adding more relevant information alongside factual error correction leads to a decline in performance. Additionally, we find that when no relevant documents are retrieved during the reasoning process, the correction model tends to append statements such as ``\textit{The retrieved documents do not provide any useful information...}'' to the summary. This can mislead the reasoning model into assuming that the problem is unsolvable, causing it to halt exploration and directly output an incorrect answer with no response, thereby reducing system robustness.
Therefore, it is more effective for CheckRLM to minimally revise the reasoning chain rather than injecting summarized retrieved information directly into it.

\subsection{Effect of Different Retrievers}
\label{ssec: Effect of different retrievers}
The Table~\ref{tab: app_retrievers} presents the impact of two different retrievers, the sparse retriever BM25 and the dense retriever bge-large-en-v1.5, on the CheckRLM framework. It can be observed that bge-large-en-v1.5 yields better performance than BM25. Furthermore, CheckRLM significantly outperforms both Direct Reasoning and Vanilla RAG across both retrieval settings, demonstrating the robustness of our proposed method.

\subsection{Cost-Efficient Reasoning Results}
\label{ssec:Analysis of Cost-Efficient Reasoning Results}
Table~\ref{tab: Cost-70B} and Table~\ref{tab: Cost-14B} present more results of Cost-Efficient Reasoning.

\begin{figure}[t]
    \centering
  \includegraphics[width=0.9\columnwidth]{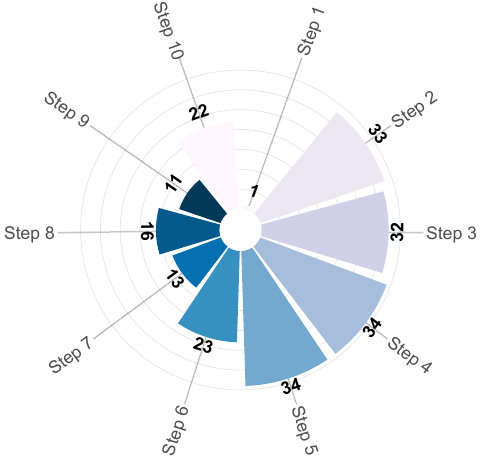}
  \caption{\textbf{Correction step distribution in 2WikiMQA.}}
  \label{fig-app:2WikiMQA-70B-flower}
\end{figure}
\begin{figure}[htbp]
    \centering
  \includegraphics[width=0.9\columnwidth]{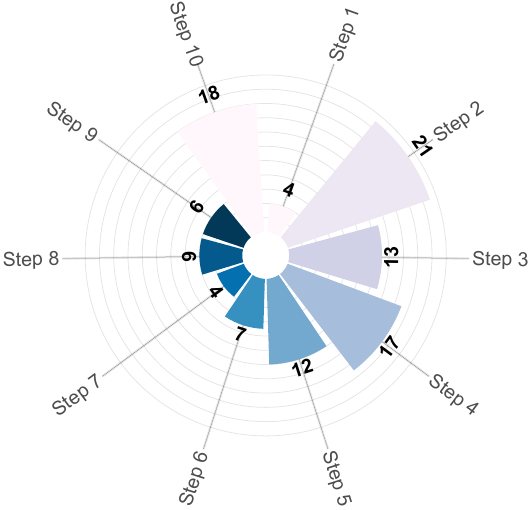}
  \caption{\textbf{Correction step distribution in MuSiQue.}}
  \label{fig-app:MusiQue-70B-flower}
\end{figure}
\subsection{Correction Step Distribution}
We use QwQ-32B as the reasoning model and Llama-3.3-70B-Instruct as the recognition and correction model.
Figures~\ref{fig-app:2WikiMQA-70B-flower} and~\ref{fig-app:MusiQue-70B-flower} present the distribution of correction steps for CheckRLM on 2WikiMQA and MuSiQue. 

Across both datasets, these distributions exhibit a strong concentration on early checking steps, particularly Step~2-5, indicating that most factual inconsistencies are corrected promptly after they are introduced. 
This pattern suggests that CheckRLM effectively intercepts errors at an early stage of the reasoning process, preventing erroneous knowledge from being propagated to subsequent reasoning steps.
Notably, deeper correction steps occur only rarely, even on MuSiQue, which involves more complex and intertwined multi-hop reasoning. 
This observation implies that the checking mechanism does not degenerate into excessive late-stage corrections, but instead resolves most errors before they accumulate and compound. 
Overall, these results highlight that the effectiveness of CheckRLM stems from early and timely correction, which plays a critical role in mitigating errors accumulated in long-step reasoning.

\section{Case Study}
Tables~\ref{tab:case_study_1},~\ref{tab:case_study_2},~\ref{tab:case_study_3} present three cases based on the same question, comparing the outputs of Direct Reasoning, CheckRLM with Post-reasoning Check, and CheckRLM with In-reasoning Check. 

Given the question ``\textit{When was the director of the film Slap Her... She's French born?}'', the correct reasoning process should first identify the director's name and then infer the birth date. 
RLM Direct Reasoning incorrectly identifies the director's name, and this error propagates to the subsequent step of determining the birth date, leading to an incorrect final answer.
Both Vanilla RAG and CheckRLM Post‑reasoning Check correctly identify the director's name; however,  they also produce an incorrect final answer due to the absence of retrieved document segments with the director's birth date.

In contrast, CheckRLM via In‑reasoning Check corrects the director's name from ``\textit{David Zucker}'' to the correct ``\textit{Melanie Mayron}'' in Step 1, and it then revises Melanie Mayron's birth date from the erroneous ``\textit{July 21, 1968}'' to the correct ``\textit{October 20, 1952}'' in Step 2, ultimately arriving at the right answer.
This illustrates the necessity of timely error correction and demonstrates the effectiveness of CheckRLM and its In‑reasoning Check strategy. Moreover, we observe that the correction model can supplement relevant knowledge pertinent to the question sometimes while correcting erroneous information, significantly streamlining the reasoning process and enabling more efficient inference.

\begin{table*}[ht]
  \centering
    \setlength{\tabcolsep}{4.5pt}
    \begin{tabular}{lccccccccccc}
    \toprule
    \multirow{3}[3]{*}{\textbf{Method}} & \multicolumn{8}{c}{\textbf{Multi-hop}} &  \multicolumn{2}{c}{\textbf{Short-form}} & \multicolumn{1}{c}{} \\
    \cmidrule(lr){2-9}\cmidrule(lr){10-11}
    & \multicolumn{2}{c}{\textbf{HotpotQA}} & \multicolumn{2}{c}{\textbf{2WikiMQA}} & \multicolumn{2}{c}{\textbf{MuSiQue}} & \multicolumn{2}{c}{\textbf{IIRC}} & \multicolumn{2}{c}{\textbf{SimpleQA}} & \textbf{Avg.} \\
    \cmidrule(lr){2-3}\cmidrule(lr){4-5}\cmidrule(lr){6-7}\cmidrule(lr){8-9}\cmidrule(lr){10-11}\cmidrule(lr){11-11}
    & f1 & em & f1 & em & f1 & em & f1 & em & f1 & em & \\
    \midrule
    \multicolumn{12}{l}{\textit{QwQ-32B}} \\
    \hdashline
    Direct Reasoning & 38.4 & 29.0 & 34.6 & 28.4 & 18.5 & 7.4 & 24.8 & 21.4 & 10.5 & 4.6 & 21.8 \\
    Vanilla RAG & 52.7 & 42.2 & 46.4 & 42.6 & 19.3 & 10.0 & 25.0 & 22.0 & 31.4 & 24.2 & 31.6\\
    \multicolumn{12}{l}{\textbf{\textit{CheckRLM}}} \\
    - Check\texttt{@}Qwen-3-8B & 59.9 & 47.4 & 59.9 & 50.6 & 29.0 & 17.8 & 29.0 & 25.4 & 33.4 & 25.2  & 37.8 \\
    - Check\texttt{@}Qwen-2.5-14B & 61.6 & 48.6 & 65.5 & 54.8 & 33.3 & 21.2 & 30.7 & 26.0 & 36.0 & 27.3 & 40.5\\
    - Check\texttt{@}Qwen-2.5-32B & 59.4 & 46.6 & 66.3 & 56.6 & 33.1 & 19.8 & 30.9 & 26.2 & 34.0 & 25.2 & 39.8\\
    - Check\texttt{@}Llama-3.3-70B & \textbf{66.3} & \textbf{52.6 } & \textbf{73.4} & \textbf{62.0} & \textbf{39.6} & \textbf{27.2} & \textbf{33.1} & \textbf{29.0} & \textbf{40.0} & \textbf{30.4} & \textbf{45.4}\\
    
    \midrule
    \multicolumn{12}{l}{\textit{Qwen3-32B}} \\
    \hdashline
    Direct Reasoning & 36.4 & 27.6 & 34.0 & 29.4 & 16.6 & 6.6 & 23.0 & 20.2 & 9.5 & 2.4 & 20.6\\
    Vanilla RAG & 51.2 & 41.0 & 44.5 & 41.8 & 19.7 & 10.4 & 22.5 & 19.8 & 29.3 & 22.8 & 30.3\\
    \multicolumn{12}{l}{\textbf{\textit{CheckRLM}}} \\
    - Check\texttt{@}Qwen-2.5-14B & 57.0 & 46.8 & 62.9 & 50.2 & 29.9 & 20.0 & 26.5 & 23.2 & 34.2 & 26.2 & 37.7\\
    - Check\texttt{@}Llama-3.3-70B & \textbf{64.2} & \textbf{52.0} & \textbf{70.8} & \textbf{60.8} & \textbf{35.6} & \textbf{25.4} & \textbf{31.6} & \textbf{27.8} & \textbf{39.1} & \textbf{30.6} & \textbf{43.8}\\
    \midrule
    \multicolumn{12}{l}{\textit{Qwen3-8B}} \\
    \hdashline
    Direct Reasoning & 29.3 & 22.2 & 31.1 & 26.6 & 12.9 & 5.2 & 21.0 & 18.2 & 7.7 & 2.8 & 17.7\\
    Vanilla RAG & 44.5 & 35.2 & 40.9 & 38.4 & 13.8 & 6.0 & 19.8 & 16.8 & 29.4 & 23.0 & 26.8\\
    \multicolumn{12}{l}{\textbf{\textit{CheckRLM}}} \\
    - Check\texttt{@}Qwen-2.5-32B & 56.3 & 45.0 & 60.0 & 50.0 & 28.2 & 18.0 & 25.0 & 22.2 & 33.0 & 26.2 & 36.4\\
    - Check\texttt{@}Llama-3.3-70B & \textbf{61.9} & \textbf{49.6} & \textbf{69.7} & \textbf{58.6} & \textbf{35.0} & \textbf{24.0} & \textbf{29.7} & \textbf{25.4} & \textbf{35.7} & \textbf{28.6} & \textbf{41.8}\\
    
    \bottomrule
    \end{tabular}%
  \caption{\textbf{Comprehensive benchmark results (\%) using different RLMs paired with various recognition and correction models.}}
  \label{tab: appendix main results}%
\end{table*}%

\begin{table*}[ht]
  \centering
    \renewcommand{\arraystretch}{1.1}
    \begin{tabular}{lcccccccccc}
    \toprule
    \multirow{3}[3]{*}{\textbf{Method}} & \multicolumn{8}{c}{\textbf{Multi-hop}} &  \multicolumn{2}{c}{\textbf{Short-form}}\\
    \cmidrule(lr){2-9}\cmidrule(lr){10-11}
    & \multicolumn{2}{c}{\textbf{HotpotQA}} & \multicolumn{2}{c}{\textbf{2WikiMQA}} & \multicolumn{2}{c}{\textbf{MuSiQue}} & \multicolumn{2}{c}{\textbf{IIRC}} & \multicolumn{2}{c}{\textbf{SimpleQA}} \\
    \cmidrule(lr){2-3}\cmidrule(lr){4-5}\cmidrule(lr){6-7}\cmidrule(lr){8-9}\cmidrule(lr){10-11}
    & f1 & em & f1 & em & f1 & em & f1 & em & f1 & em \\
    \midrule
    
    \multicolumn{11}{l}{\textit{Reasoning: QwQ-32B \& Checking: Qwen-2.5-14B}} \\
    \hdashline
    Vanilla RAG & 52.7  & 42.2  & 46.4  & 42.6  & 19.3  & 10.0  & 25.0  & 22.0  & 31.4  & 24.2 \\
    Inference Only & 61.6  & 48.6  & 65.5  & 54.8  & 33.3  & 21.2  & 30.7  & 26.0  & 36.0  & 27.3  \\
    DPO Training & \textbf{63.2} & \textbf{50.0} & \textbf{71.2} & \textbf{58.4} & \textbf{36.2} & \textbf{22.0} & \textbf{32.2} & \textbf{28.0} & \textbf{36.9} & \textbf{28.0}  \\
    
    \midrule
    
    \multicolumn{11}{l}{\textit{Reasoning: Qwen3-32B \& Checking: Qwen-2.5-14B}} \\
    \hdashline
    Vanilla RAG & 51.2  & 41.0  & 44.5  & 41.8  & 19.7  & 10.4  & 22.5  & 19.8  & 29.3  & 22.8   \\
    Inference Only & 57.0  & 46.8  & 62.9  & 50.2  & 29.9  & 20.0  & 26.5  & 23.2  & 34.2  & 26.2   \\
    DPO Training & \textbf{61.3} & \textbf{48.6} & \textbf{65.3} & \textbf{54.8} & \textbf{32.7} & \textbf{20.2} & \textbf{31.0} & \textbf{26.6} & \textbf{36.1} & \textbf{28.8}  \\

    \midrule

    \multicolumn{11}{l}{\textit{Reasoning: DeepSeek-R1-Distill-Llama-70B \& Checking: Qwen-2.5-14B}} \\
    \hdashline
    Vanilla RAG & 56.1  & 43.0  & 49.4  & 44.4  & 25.1  & 14.6  & 26.1  & 22.6  & 32.0  & 25.0 \\
    Inference Only & 60.9  & 47.4  & 64.1  & 53.4  & 32.3  & \textbf{20.8}  & 29.0  & 23.0  & 35.9  & \textbf{27.9} \\
    DPO Training & \textbf{62.7} & \textbf{49.2} & \textbf{65.7} & \textbf{54.4} & \textbf{31.7} & 20.0 & \textbf{31.2} & \textbf{26.0} & \textbf{36.8} & 27.8 \\
    
    \bottomrule
    \end{tabular}%
  \caption{\textbf{Results (\%) of performance with DPO training.}}
  \label{tab-app: dpo results}%
\end{table*}%

\begin{table*}[ht]
  \centering
    \renewcommand{\arraystretch}{1.1}
    \begin{tabular}{lcccccccccc}
    \toprule
    \multirow{3}[3]{*}{\textbf{Method}} & \multicolumn{8}{c}{\textbf{Multi-hop}} &  \multicolumn{2}{c}{\textbf{Short-form}}\\
    \cmidrule(lr){2-9}\cmidrule(lr){10-11}
    & \multicolumn{2}{c}{\textbf{HotpotQA}} & \multicolumn{2}{c}{\textbf{2WikiMQA}} & \multicolumn{2}{c}{\textbf{MuSiQue}} & \multicolumn{2}{c}{\textbf{IIRC}} & \multicolumn{2}{c}{\textbf{SimpleQA}} \\
    \cmidrule(lr){2-3}\cmidrule(lr){4-5}\cmidrule(lr){6-7}\cmidrule(lr){8-9}\cmidrule(lr){10-11}
    & f1 & em & f1 & em & f1 & em & f1 & em & f1 & em \\
    \midrule
    
    \multicolumn{11}{l}{\textit{Reasoning: QwQ-32B \& Checking: Qwen-2.5-14B}} \\
    \hdashline
    $\mathcal{D}_{\rm \text{KCR}}$ Only & 59.9  & 48.0  & 67.9  & 56.2  & 33.7  & 21.2  & 31.2  & 27.0  & 36.2  & 27.4 \\
    $\mathcal{D}_{\rm \text{KCC}}$ Only & 60.3  & 47.6  & 68.9 & 56.0  & 33.8 & 20.6  & 29.9 & 25.6  & 34.1 & 24.0  \\
    $\mathcal{D}_{\rm \text{KCR}}$+$\mathcal{D}_{\rm \text{KCC}}$ & \textbf{63.2} & \textbf{50.0} & \textbf{71.2} & \textbf{58.4} & \textbf{36.2} & \textbf{22.0} & \textbf{32.2} & \textbf{28.0} & \textbf{36.9} & \textbf{28.0}  \\
    \bottomrule
    \end{tabular}%
  \caption{\textbf{Results (\%) of performance with different training data compositions.}}
  \label{tab-app: different dpo results}%
\end{table*}%

\begin{table*}[ht]
  \centering
    \begin{tabular}{lcccccccccc}
    \toprule
    \multirow{3}[3]{*}{\textbf{Method}} & \multicolumn{8}{c}{\textbf{Multi-hop}} &  \multicolumn{2}{c}{\textbf{Short-form}}\\
    \cmidrule(lr){2-9}\cmidrule(lr){10-11}
    & \multicolumn{2}{c}{\textbf{HotpotQA}} & \multicolumn{2}{c}{\textbf{2WikiMQA}} & \multicolumn{2}{c}{\textbf{MuSiQue}} & \multicolumn{2}{c}{\textbf{IIRC}} & \multicolumn{2}{c}{\textbf{SimpleQA}} \\
    \cmidrule(lr){2-3}\cmidrule(lr){4-5}\cmidrule(lr){6-7}\cmidrule(lr){8-9}\cmidrule(lr){10-11}
    & f1 & em & f1 & em & f1 & em & f1 & em & f1 & em \\
    \midrule
    
    \multicolumn{11}{l}{\textit{Reasoning: Qwen3-8B \& Checking: Qwen2.5-32B}} \\
    \hdashline
    \textbf{CheckRLM} & \textbf{56.3}  & \textbf{45.0}  & \textbf{60.0}  & \textbf{50.0}  & \textbf{28.2}  & \textbf{18.0}  & \textbf{25.0}  & \textbf{22.2}  & \textbf{33.0}  & \textbf{26.2}  \\
    \,\,\,\,- w/o Verification & 45.3  & 36.8  & 54.4  & 47.0  & 20.1  & 10.8  & 24.2  & 20.8  & 24.3  & 17.0  \\
    \,\,\,\,- w/o Refinement & 50.2  & 39.8  & 56.0  & 48.2  & 26.2  & 15.2  & 24.1  & 20.4  & 32.3 & 24.8 \\
    
    \midrule
    
    \multicolumn{11}{l}{\textit{Reasoning: QwQ-32B \& Checking: Qwen2.5-14B}} \\
    \hdashline
    \multicolumn{1}{l}{\textbf{CheckRLM}} & \textbf{61.6} & \textbf{48.6} & \textbf{65.5} & \textbf{54.8} & \textbf{33.3} & \textbf{21.2} & \textbf{30.7} & \textbf{26.0} & \textbf{36.0} & \textbf{27.3} \\
    \,\,\,\,- w/o Verification & 52.2  & 40.2  & 62.0  & 54.2  & 30.2  & 18.6  & 29.1  & 25.2  & 30.6  & 21.6 \\
    \,\,\,\,- w/o Refinement & 58.0  & 45.4  & 64.2  & 53.0  & 29.7 & 18.6 & 28.2  & 23.0  & 33.3 & 25.8 \\

    \midrule
    
    \multicolumn{11}{l}{\textit{Reasoning: QwQ-32B \& Checking: Llama-3.3-70B}} \\
    \hdashline
    \multicolumn{1}{l}{\textbf{CheckRLM}} & \textbf{66.3} & \textbf{52.6} & \textbf{73.4} & \textbf{62.0} & \textbf{39.6} & \textbf{27.2} & \textbf{33.1} & \textbf{29.0} & \textbf{40.0} & \textbf{30.4} \\
    \,\,\,\,- w/o Verification & 62.9  & 50.4  & 72.6  & 60.8  & 39.4  & 26.0  & 33.0  & 28.6  & 37.7  & 28.4 \\
    \,\,\,\,- w/o Refinement & 64.9  & 52.4  & 68.0  & 56.8  & 37.5 & 23.6 & 32.7  & 28.4  & 36.8 & 29.0 \\
    
    \bottomrule
    \end{tabular}%
  \caption{\textbf{All results (\%) of constrained ablation study.}}
  \label{tab: ablation}%
\end{table*}%

\begin{table*}[ht]
  \centering
    \begin{tabular}{lcccccccccc}
    \toprule
    \multirow{3}[3]{*}{\textbf{Method}} & \multicolumn{8}{c}{\textbf{Multi-hop}} &  \multicolumn{2}{c}{\textbf{Short-form}}\\
    \cmidrule(lr){2-9}\cmidrule(lr){10-11}
    & \multicolumn{2}{c}{\textbf{HotpotQA}} & \multicolumn{2}{c}{\textbf{2WikiMQA}} & \multicolumn{2}{c}{\textbf{MuSiQue}} & \multicolumn{2}{c}{\textbf{IIRC}} & \multicolumn{2}{c}{\textbf{SimpleQA}} \\
    \cmidrule(lr){2-3}\cmidrule(lr){4-5}\cmidrule(lr){6-7}\cmidrule(lr){8-9}\cmidrule(lr){10-11}
    & f1 & em & f1 & em & f1 & em & f1 & em & f1 & em \\
    \midrule
    \multicolumn{11}{l}{{\textit{Reasoning: QwQ-32B \& Checking: Llama-3.3-70B}}} \\
    \hdashline
    \textbf{CheckRLM} & \textbf{66.3} & \textbf{52.6 } & \textbf{73.4} & \textbf{62.0} & 39.6 & \textbf{27.2} & \textbf{33.1} & \textbf{29.0} & \textbf{40.0} & \textbf{30.4} \\
    \,\,\,\, - w/ Summary & 66.2 & 52.0 & 71.5 & 59.6 & \textbf{39.8} & 25.2 & 33.0 & 28.0 & 38.5 & 29.8 \\
    \bottomrule
    \end{tabular}%
  \caption{\textbf{Comparison (\%) of CheckRLM with and without retrieved summary injection.} ``w/ Summary'' denotes injecting summarized retrieved information into the reasoning chain during correction.}
  \label{tab-app: Summary}%
\end{table*}%

\begin{table*}[ht]
  \centering
    \begin{tabular}{lcccccccccc}
    \toprule
    \multirow{3}[3]{*}{\textbf{Method}} & \multicolumn{8}{c}{\textbf{Multi-hop}} &  \multicolumn{2}{c}{\textbf{Short-form}}\\
    \cmidrule(lr){2-9}\cmidrule(lr){10-11}
    & \multicolumn{2}{c}{\textbf{HotpotQA}} & \multicolumn{2}{c}{\textbf{2WikiMQA}} & \multicolumn{2}{c}{\textbf{MuSiQue}} & \multicolumn{2}{c}{\textbf{IIRC}} & \multicolumn{2}{c}{\textbf{SimpleQA}} \\
    \cmidrule(lr){2-3}\cmidrule(lr){4-5}\cmidrule(lr){6-7}\cmidrule(lr){8-9}\cmidrule(lr){10-11}
    & f1 & em & f1 & em & f1 & em & f1 & em & f1 & em \\
    \midrule
    
    \multicolumn{11}{l}{\textit{Reasoning: Qwen3-8B \& Checking: Qwen2.5-32B}} \\
    \hdashline
    Post‑reasoning Check & 46.8 & 36.8  & 49.1 & 42.6  & 17.9 & 8.2 & 24.3 & 20.4 & 30.7 & 22.6 \\
    In‑reasoning Check & \textbf{56.3} & \textbf{45.0 } & \textbf{60.0} & \textbf{50.0} & \textbf{28.2} & \textbf{18.0} & \textbf{25.0} & \textbf{22.2} & \textbf{33.0} & \textbf{26.2} \\
    \textcolor[rgb]{ .129,  .361,  .596}{\% improv.} & \textcolor[rgb]{ .129,  .361,  .596}{9.5} & \textcolor[rgb]{ .129,  .361,  .596}{8.2} & \textcolor[rgb]{ .129,  .361,  .596}{10.9} & \textcolor[rgb]{ .129,  .361,  .596}{7.4} & \textcolor[rgb]{ .129,  .361,  .596}{10.3} & \textcolor[rgb]{ .129,  .361,  .596}{9.8} & \textcolor[rgb]{ .129,  .361,  .596}{0.7} & \textcolor[rgb]{ .129,  .361,  .596}{1.8} & \textcolor[rgb]{ .129,  .361,  .596}{2.3} & \textcolor[rgb]{ .129,  .361,  .596}{3.6} \\
    
    \midrule
    
    \multicolumn{11}{l}{\textit{Reasoning: QwQ-32B \& Checking: Qwen2.5-14B}} \\
    \hdashline
    Post‑reasoning Check & 46.7 & 35.8 & 47.6 & 39.6  & 24.3 & 13.0 & 28.3 & 23.6 & 26.3 & 18.6\\
    In‑reasoning Check & \textbf{61.6} & \textbf{48.6 } & \textbf{65.5} & \textbf{54.8} & \textbf{33.3} & \textbf{21.2} & \textbf{30.7} & \textbf{26.0} & \textbf{36.0} & \textbf{27.3} \\
    \textcolor[rgb]{ .129,  .361,  .596}{\% improv.} & \textcolor[rgb]{ .129,  .361,  .596}{14.9} & \textcolor[rgb]{ .129,  .361,  .596}{12.8} & \textcolor[rgb]{ .129,  .361,  .596}{17.9} & \textcolor[rgb]{ .129,  .361,  .596}{15.2} & \textcolor[rgb]{ .129,  .361,  .596}{9.0} & \textcolor[rgb]{ .129,  .361,  .596}{8.2} & \textcolor[rgb]{ .129,  .361,  .596}{2.4} & \textcolor[rgb]{ .129,  .361,  .596}{2.4} & \textcolor[rgb]{ .129,  .361,  .596}{9.7} & \textcolor[rgb]{ .129,  .361,  .596}{8.7} \\

    \midrule
    
    \multicolumn{11}{l}{\textit{Reasoning: QwQ-32B \& Checking: Llama-3.3-70B}} \\
    \hdashline
    Post‑reasoning Check & 58.8 & 47.0 & 60.2 & 52.2  & 31.5 & 19.4 & 32.2 & 27.6 & 38.3 & 29.0\\
    In‑reasoning Check & \textbf{66.3} & \textbf{52.6 } & \textbf{73.4} & \textbf{62.0} & \textbf{39.6} & \textbf{27.2} & \textbf{33.1} & \textbf{29.0} & \textbf{40.0} & \textbf{30.4} \\
    \textcolor[rgb]{ .129,  .361,  .596}{\% improv.} & \textcolor[rgb]{ .129,  .361,  .596}{7.5} & \textcolor[rgb]{ .129,  .361,  .596}{5.6} & \textcolor[rgb]{ .129,  .361,  .596}{13.2} & \textcolor[rgb]{ .129,  .361,  .596}{9.8} & \textcolor[rgb]{ .129,  .361,  .596}{8.1} & \textcolor[rgb]{ .129,  .361,  .596}{7.8} & \textcolor[rgb]{ .129,  .361,  .596}{0.9} & \textcolor[rgb]{ .129,  .361,  .596}{1.4} & \textcolor[rgb]{ .129,  .361,  .596}{1.7} & \textcolor[rgb]{ .129,  .361,  .596}{1.4} \\
    
    \bottomrule
    \end{tabular}%
  \caption{\textbf{Post‑reasoning Check vs. In‑reasoning Check.} We compare the performance of CheckRLM with Post‑reasoning Check and In‑reasoning Check.}
  \label{tab-app: local_global}%
\end{table*}%

\begin{table*}[ht]
  \centering
    \begin{tabular}{lcccccccc}
    \toprule
    \multirow{3}[3]{*}{\textbf{Method}} & \multicolumn{8}{c}{\textbf{Multi-hop}} \\
    \cmidrule(lr){2-9}
    & \multicolumn{2}{c}{\textbf{HotpotQA}} & \multicolumn{2}{c}{\textbf{2WikiMQA}} & \multicolumn{2}{c}{\textbf{MuSiQue}} & \multicolumn{2}{c}{\textbf{IIRC}} \\
    \cmidrule(lr){2-3}\cmidrule(lr){4-5}\cmidrule(lr){6-7}\cmidrule(lr){8-9}
    & f1 & em & f1 & em & f1 & em & f1 & em \\
    \midrule
    \multicolumn{9}{l}{\textit{QwQ-32B}} \\
    \hdashline
    Direct Reasoning & 38.4 & 29.0 & 34.6 & 28.4 & 18.5 & 7.4 & 24.8 & 21.4 \\
    Vanilla RAG & 52.7 & 42.2 & 46.4 & 42.6 & 19.3 & 10.0 & 25.0 & 22.0\\
    \multicolumn{9}{l}{\textbf{\textit{CheckRLM}}} \\
    BM25 & \textbf{66.3} & \textbf{52.6} & 73.4 & 62.0 & 39.6 & \textbf{27.2} & 33.1 & 29.0 \\
    bge-large-en-v1.5 & 65.2 & 51.8 & \textbf{74.3} & \textbf{63.8} & \textbf{40.1} & 26.0 & \textbf{35.7} & \textbf{31.0} \\
    
    \midrule
    \multicolumn{9}{l}{\textit{Qwen3-32B}} \\
    \hdashline
    Direct Reasoning & 36.4 & 27.6 & 34.0 & 29.4 & 16.6 & 6.6 & 23.0 & 20.2\\
    Vanilla RAG & 51.2 & 41.0 & 44.5 & 41.8 & 19.7 & 10.4 & 22.5 & 19.8 \\
    \multicolumn{9}{l}{\textbf{\textit{CheckRLM}}} \\
    BM25 & 64.2 & 52.0 & 70.8 & 60.8 & 35.6 & 25.4 & 31.6 & 27.8 \\
    bge-large-en-v1.5 & \textbf{67.7} & \textbf{54.2} & \textbf{74.2} & \textbf{63.4} & \textbf{39.6} & \textbf{26.0} & \textbf{34.3} & \textbf{29.8} \\

    \midrule
    \multicolumn{9}{l}{\textit{Qwen3-8B}} \\
    \hdashline
    Direct Reasoning & 29.3 & 22.2 & 31.1 & 26.6 & 12.9 & 5.2 & 21.0 & 18.2 \\
    Vanilla RAG & 44.5 & 35.2 & 40.9 & 38.4 & 13.8 & 6.0 & 19.8 & 16.8 \\
    \multicolumn{9}{l}{\textbf{\textit{CheckRLM}}} \\
    BM25 & 61.9 & 49.6 & 69.7 & 58.6 & 35.0 & 24.0 & 29.7 & 25.4 \\
    bge-large-en-v1.5 & \textbf{64.9} & \textbf{51.8} & \textbf{74.7} & \textbf{64.6} & \textbf{36.1} & \textbf{24.8} & \textbf{35.2} & \textbf{30.6} \\
    
    \bottomrule
    \end{tabular}%
  \caption{\textbf{Results (\%) of different retrievers.} We present the results of CheckRLM using Llama-3.3-70B-Instruct as the recognition and correction model. }
  \label{tab: app_retrievers}%
\end{table*}%

\begin{table*}[ht]
  \centering
    \begin{tabular}{lccc}
    \hline
    \textbf{Method} & \textbf{\# Tokens\,\,($\downarrow$)} & \textbf{Time (s)\,\,($\downarrow$)} & \textbf{f1 (\%)}\,\,($\uparrow$) \\
    \hline
    Direct Reasoning & 910.9  & 1.8   & 34.6 \\
    Vanilla RAG & 1684.0  & 4.0   & 46.4 \\
    Search-o1 & 2008.5 & 3.9 & 71.4 \\
    \hdashline
    Post‑reasoning Check & 1487.2  & 2.9   & 60.2 \\
    In‑reasoning Check & 1364.1  & 3.3   & 73.4 \\
    \hline
    \end{tabular}%
  \caption{\textbf{Cost-Efficient reasoning results (\%).} We conduct experiments on the 2WikiMQA dataset, using QwQ-32B as the reasoning model and Llama-3.3-70B-Instruct as the recognition and correction model.}
  \label{tab: Cost-70B}%
\end{table*}%

\begin{table*}[ht]
  \centering
    \begin{tabular}{lccc}
    \hline
    \textbf{Method} & \textbf{\# Tokens\,\,($\downarrow$)} & \textbf{Time (s)\,\,($\downarrow$)} & \textbf{f1 (\%)}\,\,($\uparrow$) \\
    \hline
    Direct Reasoning & 910.9  & 1.8   & 34.6 \\
    Vanilla RAG & 1684.0  & 4.0   & 46.4 \\
    Search-o1 & 2008.5 & 3.9 & 71.4 \\
    \hdashline
    Post‑reasoning Check & 1496.7  & 4.2   & 47.6 \\
    In‑reasoning Check & 1082.8  & 2.0   & 65.5 \\
    \hline
    \end{tabular}%
  \caption{\textbf{Cost-Efficient reasoning results (\%).} We conduct experiments on the 2WikiMQA dataset, using QwQ-32B as the reasoning model and Qwen2.5-14B-Instruct as the recognition and correction model.}
  \label{tab: Cost-14B}%
\end{table*}%


\begin{table*}[t]
\centering

\begin{tabular}{>{\raggedright\arraybackslash}p{0.9\linewidth}}
\toprule

\textbf{Query:} When was the director of film Slap Her... She'S French born? \\
\textbf{Golden Reasoning Path:} Slap Her… She's French director \textcolor{OliveGreen}{\textbf{Melanie Mayron}} -> Melanie Mayron date of birth \textcolor{OliveGreen}{\textbf{October 20, 1952}} \\
\textbf{Answer:} October 20, 1952. \\

\hdashline
\textbf{Direct Reasoning Output:} \\

\colorbox{yellow!20}{<think>} Okay, I need to find out when the director of the film ``Slap Her... She's French'' was born. First, I should figure out who the director of that movie is. Let me recall... I think the movie is from the 1980s. The title sounds familiar. Maybe it's directed by someone like \textcolor{BrickRed}{\textbf{Jim Abrahams}}? Wait, Jim Abrahams was part of the team that did the Police Academy movies. Let me check that.

Alternatively, maybe it's \textcolor{BrickRed}{\textbf{Mel Brooks}}? No, Mel Brooks did Blazing Saddles and Young Frankenstein. Wait, Slap Her... She's French, that sounds like a cult movie. Let me think again. Oh, right, the director is Jim Abrahams. Wait, but I'm not 100\% sure. Let me confirm. The movie came out in 1975, I believe. The director is Jim Abrahams, but I should make sure. 

Alternatively, maybe it's his brother, David, or one of the other brothers? Wait, Jim Abrahams, David Zucker, and Jerry Zucker? No, maybe not. Let me think again. The film ``Slap Her... She's French'' is a 1975 comedy. The director is indeed Jim Abrahams. Now, when was Jim Abrahams born? 

\textcolor{BrickRed}{\textbf{Jim Abrahams was born on February 25, 1940.}} Wait, but I need to confirm that. Let me think. Jim Abrahams' birth date. I might be mixing up with his brother David. Wait, David Zucker was born in 1949. Hmm. Alternatively, maybe Jim Abrahams was born in 1940. Let me see. 

Alternatively, perhaps I can recall that he was part of the team that did the Airplane! movies, but that's more the Zucker-Abrahams team. Wait, Jim Abrahams was born in 1940. Let me check in my mind. If the movie was 1975, and he was, say, in his mid-30s then, that would make his birth year around 1940. So, I think the correct answer is February 25, 1940. So the director, Jim Abrahams, was born on that date. I'll go with that.\colorbox{yellow!20}{</think>} \\
\textbf{Answer:} \textcolor{BrickRed}{\textbf{February 25, 1940}}

\textbf{Vanilla RAG Output:} \\

\colorbox{yellow!20}{<think>} Okay, I need to find out when the director of the film ``Slap Her... She's French'' was born. Let's look at the given passages.

First, the film's title is mentioned in two passages. Both say the film is directed by \textcolor{OliveGreen}{\textbf{Melanie Mayron}} and starring Piper Perabo. The title of the film is ``Slap Her... She's French,'' which was released in 2002. The question is about the director's birth year, so I need to find information related to Melanie Mayron.

Looking at the third passage about Jane McGregor, it states she was born in 1983 and lists her roles, including ``Slap Her ... She's French'' from 2002. However, Jane McGregor is an actress, not the director. The director is Melanie Mayron, whose details aren't provided in any of the given texts. The passages don't mention Melanie Mayron's birth date. The only birth year given here is Jane McGregor's, but that's irrelevant to the director. Since the information isn't present in the provided passages, \textcolor{BrickRed}{\textbf{the answer can't be determined from the given data}}. The user's question might be expecting me to realize that the passages don't include the director's birth year. \colorbox{yellow!20}{</think>} \\ 
\textbf{Answer:} \textcolor{BrickRed}{\textbf{Unknown}} \\
\bottomrule
\caption{\textbf{Case Study I.} This case illustrates the outputs of Direct Reasoning via RLM and vanilla RAG. We use QwQ-32B as the backbone model for reasoning.}
\label{tab:case_study_1}
\end{tabular}
\end{table*}

\begin{table*}[t]
\centering

\begin{tabular}{>{\raggedright\arraybackslash}p{0.9\linewidth}}
\toprule

\textbf{Query:} When was the director of film Slap Her... She'S French born? \\
\textbf{Golden Reasoning Path:} Slap Her… She's French director \textcolor{OliveGreen}{\textbf{Melanie Mayron}} -> Melanie Mayron date of birth \textcolor{OliveGreen}{\textbf{October 20, 1952}} \\
\textbf{Answer:} October 20, 1952. \\

\hdashline
\textbf{CheckRLM Post‑reasoning Check Output:} \\
\textbf{Reasoning Output:} \colorbox{yellow!20}{<think>} Okay, I need to find out when the director of the film ``Slap Her... She's French'' was born. First, I should figure out who the director of that movie is. Let me recall... I think the movie is from the 1980s. The title sounds familiar. Maybe it's directed by someone like Jim Abrahams? Wait, Jim Abrahams was part of the team that did the Police Academy movies. Let me check that.

Alternatively, maybe it's Mel Brooks? No, Mel Brooks did Blazing Saddles and Young Frankenstein. Wait, Slap Her... She's French, that sounds like a cult movie. Let me think again. Oh, right, the director is \textcolor{BrickRed}{\textbf{Jim Abrahams}}. Wait, but I'm not 100\% sure. Let me confirm. The movie came out in 1975, I believe. The director was Jim Abrahams, who was part of the Three Amigos, along with David and Jerry. So Jim Abrahams is the director here. Now, when was he born?

Jim Abrahams was born on... I think he was born in the 1930s or 1940s. Let me recall. David and Jerry are his brothers, part of the comedy trio. Wait, Jim Abrahams was born on July 16, 1940. Let me see if that's correct. Alternatively, maybe 1941. Hmm, I might need to double-check. Wait, another way: if the movie was released in 1975, and he was in his 30s then, that would make his birth year around 1940-1945. So 1940 seems right. I think that's correct. So the answer should be \textcolor{BrickRed}{\textbf{July 16, 1940}}.\colorbox{yellow!20}{</think>} \\
\textbf{Factual Claim:} \\
1. Jim Abrahams is the director of the film 'Slap Her... She's French'.\\
2. Jim Abrahams was born on July 16, 1940. \\
\textbf{Refine Output:} \colorbox{yellow!20}{<think>} Okay, I need to find out when the director of the film ``Slap Her... She's French'' was born. First, I should figure out who the director of that movie is. Let me recall... I think the movie is from the 2000s. The title sounds familiar. Maybe it's directed by someone like Jim Abrahams? Wait, Jim Abrahams was part of the team that did the Police Academy movies. Let me check that.

Alternatively, maybe it's Mel Brooks? No, Mel Brooks did Blazing Saddles and Young Frankenstein. Wait, Slap Her... She's French, that sounds like a cult movie. Let me think again. Oh, right, the director is \textcolor{OliveGreen}{\textbf{Melanie Mayron}}. Wait, but I'm not 100\% sure. Let me confirm. The movie came out in 2002, I believe. The director was Melanie Mayron. Now, when was she born?

\textcolor{BrickRed}{\textbf{I do not have the exact birth date of Melanie Mayron from the provided documents}}. However, based on the information that the movie was released in 2002 and considering she was likely in her early career or mid-career at that time, I would need to look up her birth year elsewhere. For the purpose of this reasoning process, I will not provide a specific birth date as it is not available in the given documents.\colorbox{yellow!20}{</think>} \\
\textbf{Answer:} \textcolor{BrickRed}{\textbf{1965}}
\\

\bottomrule
\caption{\textbf{Case Study II.} This case illustrates the output of CheckRLM via Post‑reasoning Check. We use QwQ-32B as the reasoning model and Llama-3.3-70B-Instruct as the recognition and correction model.}
\label{tab:case_study_2}
\end{tabular}
\end{table*}

\begin{table*}[t]
\centering

\begin{tabular}{>{\raggedright\arraybackslash}p{0.9\linewidth}}
\toprule

\textbf{Query:} When was the director of film Slap Her... She'S French born? \\
\textbf{Golden Reasoning Path:} Slap Her… She's French director \textcolor{OliveGreen}{\textbf{Melanie Mayron}} -> Melanie Mayron date of birth \textcolor{OliveGreen}{\textbf{October 20, 1952}} \\
\textbf{Answer:} October 20, 1952. \\

\hdashline
\textbf{CheckRLM In‑reasoning Output:} \\
\textbf{Step 1:} \\
\textbf{Reasoning Output:} \colorbox{yellow!20}{<think>} Okay, I need to find out when the director of the film ``Slap Her... She's French'' was born. First, I should figure out who the director of that movie is. Let me recall... I think the movie is from the 1980s. The title sounds familiar. Maybe it's \textcolor{BrickRed}{\textbf{directed by someone like David Zucker}}? Wait, the Zucker-Abrahams-Zucker team did a lot of comedies around that time. Let me confirm. \\
\textbf{Factual Claim:} \\
1. The movie 'Slap Her... She's French' is from the 1980s. \\
2. The Zucker-Abrahams-Zucker team directed many comedies around that time. \\
\textbf{Refine Output:} \colorbox{yellow!20}{<think>} Okay, I need to find out when the director of the film ``Slap Her... She's French'' was born. First, I should figure out who the director of that movie is. Let me recall... The movie is from 2002, not the 1980s. The title sounds familiar, but it's not directed by someone like David Zucker. \textcolor{OliveGreen}{\textbf{The movie is directed by Melanie Mayron}}. Let me confirm Melanie Mayron's birth date.\\
\textbf{Step 2:} \\
\textbf{Reasoning Output:} Wait, I think \textcolor{BrickRed}{\textbf{Melanie Mayron was born on July 21, 1968}}. I should double-check that. Alternatively, maybe I'm mixing her up with someone else. Let me see... Another thought: sometimes people confuse the director with the cast. The film's cast includes actors like Adam Goldberg, but the director is definitely Melanie Mayron. \\
\textbf{Factual Claim:} \\
1. Melanie Mayron is the director of the film Slap Her... She'S French. \\
2. Melanie Mayron was born on July 21, 1968. \\
\textbf{Refine Output:} Wait, I think \textcolor{OliveGreen}{\textbf{Melanie Mayron was born on October 20, 1952}}. I should double-check that. Alternatively, maybe I'm mixing her up with someone else. Let me see... Another thought: sometimes people confuse the director with the cast. The film's cast includes actors like Piper Perabo, but the director is definitely Melanie Mayron.\\
\textbf{Step 3:} \\
\textbf{Reasoning Output:} So, Melanie Mayron's birth date is October 20, 1952. So the answer should be October 20, 1952. I'll make sure there's no conflicting info. Another check: if I remember correctly, she was born in 1952. Yeah, that seems right. So the director's birth date is 1952-10-20. \colorbox{yellow!20}{</think>}\\
\textbf{Factual Claim:} \\
1. Melanie Mayron's birth date is October 20, 1952. \\
2. The director of the film Slap Her... She'S French is Melanie Mayron. \\
\textbf{Refine Output:} So, Melanie Mayron's birth date is October 20, 1952. So the answer should be October 20, 1952. I'll make sure there's no conflicting info. Another check: if I remember correctly, she was born in 1952. Yeah, that seems right. So the director's birth date is 1952-10-20. \colorbox{yellow!20}{</think>}\\
\textbf{Answer:} \textcolor{OliveGreen}{\textbf{October 20, 1952}}\\

\bottomrule
\caption{\textbf{Case Study III.} This case illustrates the output of CheckRLM via In‑reasoning Check. We use QwQ-32B as the reasoning model and Llama-3.3-70B-Instruct as the recognition and correction model.}
\label{tab:case_study_3}
\end{tabular}
\end{table*}

\end{document}